\documentclass[review]{elsarticle}
\graphicspath{ {./figures/} }
\usepackage{hyperref}
\usepackage{float}
\usepackage{verbatim} 
\usepackage{apalike}

\usepackage{graphicx}

\usepackage{hyperref}

\usepackage{microtype}
\usepackage{graphicx}
\usepackage{subfigure}
\usepackage{booktabs} 
\usepackage{amsmath,amssymb}
\usepackage{amsthm}
\usepackage{caption}
\usepackage{multirow}
\usepackage{adjustbox}
\usepackage{upgreek}

\usepackage{amsmath,amsfonts,amssymb}
\usepackage{mathtools}
\usepackage{commath}
\usepackage{color}

\theoremstyle{definition}
\newtheorem{definition}{Definition}[section]
\newtheorem{theorem}{Theorem}[section]

\newtheorem{prop}{Proposition}[section]

\DeclareMathOperator*{\argmax}{arg\,max}
\DeclareMathOperator*{\argmin}{arg\,min}

\definecolor{todocolor}{rgb}{0.9,0.1,0.1}
\definecolor{lcolor}{rgb}{0.7,0.7,0.3}
\definecolor{qcolor}{rgb}{0,0,1}

\restylefloat{figure}
\restylefloat{table}

\journal{Expert Systems with Applications}

\biboptions{authoryear}

\begin{document}
\begin{frontmatter}

\begin{titlepage}
\begin{center}
\vspace*{1cm}

\textbf{ \large Achieving counterfactual fairness \\ with imperfect structural causal model}

\vspace{1.5cm}

Tri Dung Duong$^{a}$ (TriDung.Duong@student.uts.edu.au), Qian Li$^b$ (qli@curtin.edu.au), Guandong Xu$^a$ (Guandong.Xu@uts.edu.au) \\

\hspace{10pt}

\begin{flushleft}
\small  
$^a$ University of Technology Sydney (UTS), Australia \\
$^b$ Curtin University, Australia


\vspace{1cm}
\textbf{Corresponding Author:} \\
Guandong Xu \\
61 Broadway, Ultimo NSW 2007, Australia \\
Tel: (+61)295143788 \\
Email: Guandong.Xu@uts.edu.au

\end{flushleft}        
\end{center}
\end{titlepage}

\title{Achieving Counterfactual Fairness \\ with Imperfect Structural Causal Model}

\author[label1]{Tri Dung Duong}
\ead{TriDung.Duong@student.uts.edu.au}

\author[label1]{Qian Li}
\ead{qli@curtin.edu.au}

\author[label1]{Guandong Xu \corref{cor1}}
\ead{Guandong.Xu@uts.edu.au}

\cortext[cor1]{Corresponding author.}
\address[label1]{University of Technology Sydney (UTS), Australia}
\address[label2]{Curtin University, Australia}

\begin{abstract}
Counterfactual fairness alleviates the discrimination between the model prediction toward an individual in the actual world (observational data) and that in 
counterfactual world (i.e., what if the individual belongs to other sensitive groups). The existing studies need to pre-define the structural causal model that captures the correlations among variables for counterfactual inference; however, the underlying causal model is usually unknown and difficult to be validated in real-world scenarios. Moreover, the misspecification of the causal model potentially leads to poor performance in model prediction and thus makes unfair decisions. In this research, we propose a novel minimax game-theoretic model for counterfactual fairness that can produce accurate results meanwhile achieve a counterfactually fair decision with the relaxation of strong assumptions of structural causal models. In addition, we also theoretically prove the error bound of the proposed minimax model.
Empirical experiments on multiple real-world datasets illustrate our superior performance in both accuracy and fairness. Source code is
available at \url{https://github.com/tridungduong16/counterfactual_fairness_game_theoretic}.
\end{abstract}


\end{frontmatter}

\section{Introduction}



As machine learning (ML) is increasingly leveraged in high-stake domains such as criminal justice \cite{angwin2016machine,berk2021fairness} or credit assessment \cite{zhang2019fairness}, the concerns regarding ethical issues in designing ML algorithms have arisen recently. Fairness is one of the most important concerns to avoid discrimination in the model prediction towards an individual or a population. 
Recent years witness an increasing number of studies that have explored fairness-aware machine learning under the causal perspective \cite{nabi2018fair,kusner2017counterfactual,zhang2018fairness,chiappa2019path}. Causal models specifically \cite{pearl2009causal} provide an intuitive and powerful way of reasoning the causal effect of sensitive attribution on the final decision. Among these studies in this line of work, counterfactual fairness is a causal and individual-level fairness notion first proposed by \cite{kusner2017counterfactual}, which considers a model counterfactually fair if its predictions are identical in both a) the original world and b) the counterfactual world where an individual belongs to another demographic group.

As the first practice of counterfactual fairness,  \cite{kusner2017counterfactual} first constructs a structural causal model using prior domain knowledge. Unobserved variables are then inferred which are independent of and have no causal relationship to the sensitive attributes. 
The inferred latent variables are thereafter used as the input for the predictive models. The main limitation of the study is that the strong assumption of the causal model is required which is however hard to achieve in a real-world setting, 
especially when it comes to a large-scale dataset with a great number of features \cite{vanderweele2009concerning,peters2016causal}. 
Additionally, even if prior knowledge of causal structure is available, counterfactual fairness algorithms
involves computing counterfactuals in the true underlying structural causal model (SCM) \cite{pearl2009causality}, and thus relies on strong impractical assumptions.
Specifically, the algorithm requires complete knowledge of the true structural equations \cite{fong2013causal,bollen2013eight,pearl2012causal}. Another obstacle is that the tabular data contains both continuous and categorical data, making them difficult to be represented by the probabilistic equations. Moreover, when removing all other features and only using non-descendants of sensitive ones, there are possibly insufficient features used for model training which can degrade the model capability and significantly deteriorate the accuracy performance. 

To tackle the above limitations, we propose a novel counterfactual fairness approach with the knowledge about structural causal models is limited. In particular, we aim to minimize the sensitive information impact on model decisions, while maintaining satisfactory model accuracy. To achieve the optimal solutions that maximize the fairness-accuracy trade-offs, we propose a minimax game-theoretic approach that consists of three main components. As shown in Figure~\ref{fig:arch}, the invariant-encoder model $p_{\theta}$ learns the invariant representation that is unchangeable from sensitive attributes. After that, the fair-learning predictive model utilizes the invariant representation as the input with the purpose of not only guaranteeing the main learning tasks but also assuring the fairness aspect, while sensitive-awareness model used both the invariant representation and sensitive information that can produce the good learning performance. For theoretical proof, we provide a theoretical analysis for the generalization bound of the minimax objective functions. To illustrate the effectiveness of our proposed method, we compare our method with state-of-the-art methods on three benchmark datasets including \texttt{Law}, \texttt{Compas} and \texttt{Adult} datasets. 
The experimental results indicate that our proposed method can achieve outstanding fairness performance in comparison with other baselines. Specifically, our contributions can be summarized as follows:

\begin{itemize}
    \item We introduce a minimax game-theoretic approach to obtain the invariant-encoder model and fair-learning predictive model that can jointly produce the counterfactually fair prediction and obtain the competitive performance on both classification and regression tasks. 
    \item We prove the theoretical generalization bounds for the adversarial algorithm of the proposed minimax model. 
    \item We perform the extensive experiments on three datasets and demonstrate the effectiveness of the proposed method to achieve satisfactory fairness and accuracy. 
\end{itemize}

\section{Preliminaries}
\label{sec:pre}
In this section, we provide notations and problem statements and then review individual and counterfactual fairness notions. 


Throughout the paper, upper-cased letters $X$ and $\boldsymbol{X}$ represent the random scalars and vectors respectively, while lower-cased letters $x$ and $\boldsymbol{x}$ denote the deterministic scalars and vectors, respectively. We consider a dataset $\mathcal{D} = \{x_i, s_i, y_i\}^n_{i=1}$ consisting of $n$ instances, where $x_i \in \boldsymbol{X}$ is the normal features (e.g. age, working hours,..), $s_i \in \boldsymbol{S}$ is the sensitive feature (e.g. race and gender), and $y_i \in Y$ is the target variable regarding individuals $i$. Sensitive features specify an individual's belongings to socially salient groups (e.g. women and Asian). $H(.)$ and $I(.)$ are the corresponding Shannon entropy and mutual information \cite{cover1991entropy}, and $\mathcal{L}(.)$ is the loss function (e.g. cross-entropy for classification tasks, mean square error for regression tasks). Finally, $f_{\theta}$ represents a neural network model parameterized by $\theta$. 

\begin{figure*}[t]
  \includegraphics[width=\textwidth]{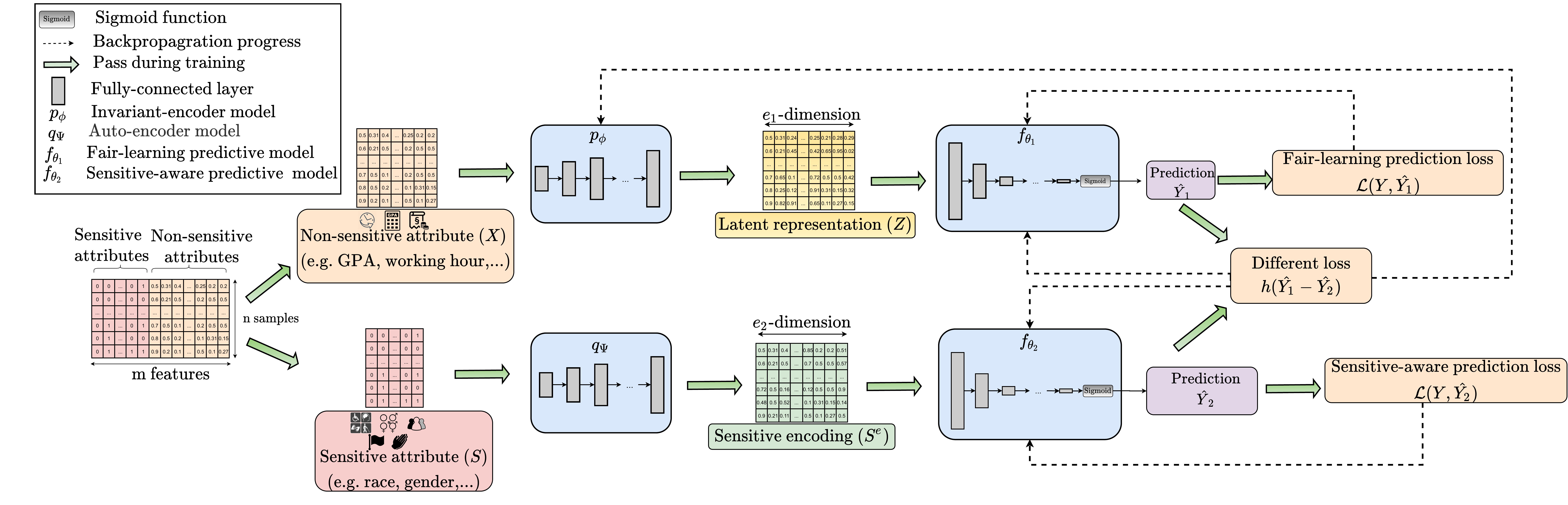}
  \caption{The framework consists of three trainable components: the invariant-encoder model $p_\phi$, fair-learning model $f_{\phi_1}$ and sensitive-awareness $f_{\phi_2}$ model.}
\label{fig:arch}
\end{figure*}

Figure~\ref{fig:arch} generally illustrates our proposed approach that consists of an invariant-encoder model ($q_\theta$) generating the invariant features, a fair-learning predictor ($f_{\phi_1}$) trained by invariant features and a sensitive-aware predictor ($f_{\phi_2}$) trained on invariant and sensitive representation. 


\begin{definition}[Counterfactual fairness \cite{kusner2017counterfactual}]
A classifier is considered as counterfactual fair given the sensitive attribute $S=s$ if:
\begin{equation}
\label{eqn:cf}
\small
    P(\hat{Y}_{S \leftarrow s} = y| X = x, S = s) = P(\hat{Y}_{S \leftarrow \hat{s}} = y| X = x, S = \hat{s}) 
\end{equation}
where $\hat{Y}$ denotes the model prediction depends on $X$ and $S$, while model prediction for intervention $S \leftarrow \hat{s}$ is denoted as $\hat{Y}_{S \leftarrow s}$. Meanwhile, $P(\hat{Y}_{S \leftarrow \hat{s}} = y| X = x, S = \hat{s})$ is the counterfactual prediction where we change the value $S=s$ to $S=\hat{s}$.

\end{definition}

The Eq.~\eqref{eqn:cf} ensures that the distribution over possible predictions is the same in both the actual world and a counterfactual world where the sensitive attribute(s) were modified while all other conditions remain unchanged.




\section{Related work}
This section focuses on the research related to our work and then highlights the main limitation of these studies.


\textbf{Individual fairness}. 
Since counterfactual fairness analyses fairness at the individual level, our work is closely related to individual fairness works. \cite{dwork2012fairness} first captures the main idea of individual fairness that two individuals having the same particular task should be treated similarly. This principle draws much attention with a plethora of studies \cite{miconi2017impossibility,biega2018equity,mukherjee2020two,sharifi2019average}. However, this concept is hard to apply in practice due to the barrier of defining the similarity regarding the individual tasks. This leads to the shift from achieving fairness decisions to defining similar tasks. Another recent study \cite{speicher2018unified} provides a unified approach to evaluate the performance of individual fairness algorithms by using a generalized entropy index that has been previously used widely in economics as a measure of income inequality in the population. Our work utilizes the generalized entropy index as the primary metric for evaluation purposes. 


\textbf{Counterfactual fairness}. 
In order to achieve fairness in the model decision, the traditional approach is the unawareness model \cite{grgic2016case} that only uses the non-sensitive attributes as the input for predictive models. This approach seems to be reasonable but neglects the biased effect of sensitive attributes on normal features. Thus, the study \cite{kusner2017counterfactual} first proposed the approach of counterfactual fairness by only using the non-descendants of the sensitive attribute for prediction tasks. They first assume the causal graph structure with the latent variables independent from the sensitive attributes. The study thereafter fits the data into the causal model and produces the posterior distribution for unobserved variables. The inferred variables are finally utilized as inputs for the predictive model. Apart from that, multi-world counterfactual fairness \cite{russell2017worlds} introduces another alternative method to deal with the uncertainty of the ground-truth causal model. They first have an assumption that there are several possible causal diagrams that represent different counterfactual worlds. Thereafter, the authors build a neural network and then use the gradient descent algorithm to minimize the difference in the predictions between the different worlds. Although this approach seems to be promising, it also needs a list of causal models to be taken into consideration. To sum up, all of the above methods require strong assumptions about causal graphs to infer the latent variables. Moreover, sensitive attributes such as race, gender, and nationality are personally intrinsic attributes and immensely influential that normally have a causal relationship to other features.

\section{Methodology}

This section illustrates our proposed method, which can achieve counterfactual fairness without the assumption of structural causal models. In summary, our proposed method aims to learn a representation along with a predictive model which together can make a counterfactual fair prediction and maintain the prediction accuracy. In summary, our proposed approach contains three main components:  1) invariant-encoder model learning the invariant representation that is unchangeable from sensitive attributes; 2) fair-learning predictive model which not only guarantees the main learning tasks but also assures the fairness aspect; 3) sensitive-awareness model that contains the sensitive information which can produce the good learning performance. Each component would be discussed in detail in Section~\ref{three}.



\begin{figure}[t]
\centering
\label{fig:graph}
  \includegraphics[width=0.4\textwidth]{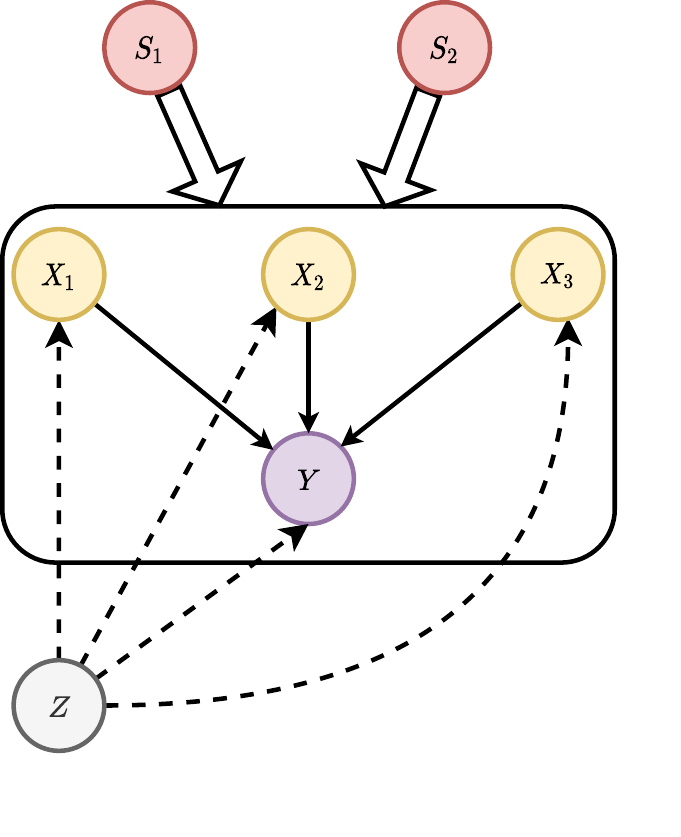}
  \caption{A structural causal model illustrates the causal relationships between different features.  $\boldsymbol{S}_1$ and $\boldsymbol{S}_2$ are sensitive features (e.g., gender or race), $\boldsymbol{X}_1$, $\boldsymbol{X}_2$ and $\boldsymbol{X}_3$ are the non-sensitive features (e.g., education or working hours), $\boldsymbol{Z}$ is a latent representation that is independent of sensitive attributes and $Y$ is the target variable. The large white arrows from $\boldsymbol{S}_1$ and $\boldsymbol{S}_2$ represent that $\boldsymbol{S}_1$ and $\boldsymbol{S}_2$ have the causal effects to every variables ($\boldsymbol{X}_1$, $\boldsymbol{X}_2$, $\boldsymbol{X}_3$) and target variable ($Y$) contained in the box. }
  \label{fig:graph}
\end{figure}

\subsection{Motivation}


We first consider an example of probabilistic graphical model in Figure~\ref{fig:graph}, we have two sensitive features: $\boldsymbol{S_1}$ and $\boldsymbol{S}_2$,
non-sensitive features $\boldsymbol{X}_1$, $\boldsymbol{X}_2$, and $\boldsymbol{X}_3$, and the target variable $Y$. We want to find a latent representation $\boldsymbol{Z}$ that is independent of the sensitive attributes. Producing the latent representation that is invariant across different sensitive attributes is a challenging task. However, we can handle this challenge if we have information about sensitive attributes. We make an assumption that the probability $p(Y|\boldsymbol{Z})$ remains the same across different sensitive attributes $\boldsymbol{S}_1$ and $\boldsymbol{S}_2$ because $\boldsymbol{Z}$ has the direct causal relationship to target variable $Y$ and also does not rely on $\boldsymbol{S}_1$ and $\boldsymbol{S}_2$. Meanwhile, other remaining features $\boldsymbol{X}_1$, $\boldsymbol{X}_2$ and $\boldsymbol{X}_3$ are causally influenced by $\boldsymbol{S}_1$ and $\boldsymbol{S}_2$; thus, the probability of $p(Y|\boldsymbol{X}_1)$, $p(Y|\boldsymbol{X}_2)$ and $p(Y|\boldsymbol{X}_3)$ will change if we change the value of $\boldsymbol{S}_1$ and $\boldsymbol{S}_2$. 




In general, our main purpose is to find a representation ($\boldsymbol{Z}$) that is invariant across different sensitive attributes. We want to design an invariant-encoder model $p_{\theta}: \boldsymbol{X} \rightarrow \boldsymbol{Z}$ that learns the representation ($\boldsymbol{Z}$) from the input ($\boldsymbol{X}$). An ideal invariant representation should satisfy the following conditions.
\begin{equation} 
\small
\label{eqn:overall}
    Y \perp do(\boldsymbol{S}) | \boldsymbol{Z} \iff H(Y|\boldsymbol{Z},do(\boldsymbol{S})) = H(Y|\boldsymbol{Z})
\end{equation}

where $\perp$ denotes probabilistic independence. $Y$ is independent
of $do(\boldsymbol{S})$ only when conditioned on latent representation $\boldsymbol{Z}$ and we call this variance property. Moreover, the Eq.~\eqref{eqn:overall} means that the representation ($\boldsymbol{Z}$) is unchangeable, and sensitive attributes ($\boldsymbol{S}$) do not provide extra information to predict target variables ($Y$). This means that making an intervention on sensitive attribute ($\boldsymbol{S}$) does not lead to changes in the model prediction ($Y$).





\subsection{Three-player model for invariant fairness}
\label{three}
Our proposed framework is illustrated in Figure~\ref{fig:arch} which describes the training and backpropagation process as well as inputs and different components. In general, the proposed framework has three main trainable models including an invariant-encoder model ($q_\theta$) that generates the invariant features, a fair-learning predictor ($f_{\phi_1}$) that predicts outcomes based on invariant features and a sensitive-aware predictor ($f_{\phi_2}$) that predicts outcomes based on both invariant and sensitive representation. The framework also includes a pre-trained auto-encoder model ($q_{\uppsi}$) that produces latent representation from sensitive features. 
For clarity, we will briefly describe the auto-encoder model and then present the two predictors followed by the invariant-encoder model.

\textbf{Auto-encoder model. }
 Sensitive attributes ($\boldsymbol{S}$) are in the categorical form and discrete values, which is hard to utilize in neural networks. Therefore, we construct the auto-encoder model ($q_{\uppsi}$) with the purpose of 1) converting discrete values to continuous form which is more suitable to the complicated models, 2) capturing the intrinsic relationship between categorical groups, and flexibly control the dimensional number of embedding vector. The auto-encoder model ($q_{\uppsi}$)\cite{ng2011sparse} is trained beforehand by using all of the features as the input. The encoder-decoder architecture with an embedding layer is used that aims to project sensitive attributes $\boldsymbol{S}$ onto an $e$-dimensional latent space $\mathcal{R}^e$ ($q_{\uppsi}: \boldsymbol{S} \rightarrow \boldsymbol{S}^e$). The latent representation of sensitive attributes ($\boldsymbol{S}^e$) would be thereafter utilized to be injected into the sensitive-aware predictor later.


\textbf{Two predictors.} The fair-learning predictor $f_{\phi_1}: \boldsymbol{Z} \rightarrow Y$ that predicts target variable ($Y$) from the latent representation ($\boldsymbol{Z}$). Meanwhile, the sensitive-aware predictive model $f_{\phi_2}:(\boldsymbol{Z},\boldsymbol{S}^e) \rightarrow Y$ makes a prediction ($Y$) from latent representation ($\boldsymbol{Z}$) and sensitive attributes information ($\boldsymbol{S}^e$). The only difference between them is that the sensitive-aware predictor can access to sensitive information, while the fair-learning one only uses the invariant representation. The loss functions for the fair-learning and sensitive-aware predictive model are $\mathcal{L}(Y, f_{\phi_1}(\boldsymbol{Z}))$ and 
$\mathcal{L}(Y, f_{\phi_2}(\boldsymbol{Z}, \boldsymbol{S}^e))$, respectively. Thus, the optimal solutions for both of them can be defined
as follows:
\begin{equation}
\label{eqn:pred1}
\small
    \phi_i^* = \argmin_{\phi_1} \mathbb{E}[\mathcal{L}(Y, f_{\phi_1}(\boldsymbol{Z}))]
\end{equation}
\begin{equation}
\label{eqn:pred2}
\small
    \phi_2^* = \argmin_{\phi_2} \mathbb{E}[\mathcal{L}(Y, f_{\phi_2}(\boldsymbol{Z}, \boldsymbol{S}^e))]
\end{equation}


\textbf{Invariant-encoder model.} The invariant-encoder model $q_{\theta}: \boldsymbol{X} \rightarrow \boldsymbol{Z}$ learns the representation ($\boldsymbol{Z}$) from the input ($\boldsymbol{X}$). The aims of the invariant-encoder model are first to optimize the fair-learning predictive model, and then to minimize the gap between the predictions of two predictive models. This allows to ensure the model accuracy in the prediction task and excludes sensitive information in model decisions. Therefore, the learning objective for the invariant-encoder model is:
\begin{equation}
\label{eqn:encoder}
\small
    \theta^* = \argmin_{\theta} \mathcal{L}(Y, f_{\phi^*_1}(\boldsymbol{Z})) + \lambda h(f_{\phi^*_1}(\boldsymbol{Z}) - f_{\phi^*_2}(\boldsymbol{Z},\boldsymbol{S}^e))
\end{equation}
where $h(t)$ is a strictly monotonic function that increases when $t > 0$, and decreases when $t < 0$. 

\textbf{Objective function and training.} By combining learning objectives from Eq.~\eqref{eqn:pred1}, ~\eqref{eqn:pred2} and~\eqref{eqn:encoder}, we can produce the final objective function in the minimax form Eq.~\eqref{eqn:objective} with the latent representation $\boldsymbol{Z} = q_{\theta}(\boldsymbol{X})$. Overall, the loss function represents the minimax game where the invariant-encoder model plays cooperative games with the fair-learning predictor and adversarial games with the sensitive-aware predictor. The objective functions first aims to minimize the prediction of function $f_{\phi_1}$ and make the gap between $f_{\phi_2}$ and $f_{\phi_1}$ as small as possible. 

\begin{equation}
\small
\label{eqn:objective}
     \argmin_{\theta, \phi_1} \argmax_{\phi_2} \mathcal{L}(Y,f_{\phi_1}(\boldsymbol{Z})) + \lambda h(f_{\phi_1}(\boldsymbol{Z}) - f_{\phi_2}(\boldsymbol{Z},\boldsymbol{S}^e))
\end{equation}

Regarding the training process for three models, the loss functions corresponding to each model are first calculated. 
We thereafter update each model by descending stochastic gradients regarding invariant-encoder model, fair-learning predictor and ascending stochastic gradient of sensitive-aware predictor. We perform updating procedure with a number of steps, and only one model is updated for each step. In our experiments, we used Adam optimization algorithm \cite{kingma2014adam} to optimize~\eqref{eqn:objective}.

\section{Theoretical analysis}
This section provides the generalization bound for our proposed method under the minimax setting.
Remember we consider the local minimax empirical risk minimization problem
\begin{equation}
\small 
\underset{\theta, \phi_{1}}{\min }\, \underset{\phi_{2}}{\max } \,\mathbb{E}[\mathcal{L}\left(Y, f_{\phi_{1}}(\boldsymbol{Z})\right)]
\label{eq:obj}
\end{equation}
By applying a duality argument, we reformulate the dual problem via the probability of sensitive attributes.
Let $P^{*}$ be the ideal fair sample distribution corresponding to $P/Q_0$, according to the underlying exposure mechanism $Q_0$ and data distribution $P$.  We choose the Wasserstein distance to investigate how to transport from the observed data distribution
to an ideal data distribution that is independent of the sensitive attributes. The reason is that unlike the Kullback-Leibler divergence, the Wasserstein metric is a true probability metric and considers both the probability of and the distance between various outcome events. Wasserstein distance provides a meaningful and smooth representation of the distance between distributions.The Wasserstein Distance is furthermore to measure distances between probability distributions on a given metric space. The use of the Wasserstein distance is motivated because this distance is defined and computable even between distributions with
disjoint supports. 


\begin{definition}[Wasserstein Distance]
The Wasserstein distance for our problem is defined as:
\begin{equation}
\small
W_{c}\left(\hat{P}, P^{*}\right)=
    \inf _{\gamma \in \Pi\left(\hat{P}, P^{*}\right)} \mathbb{E}_{\left((\mathbf{x}, \mathbf{z}, y),\left(\mathbf{x}^{\prime}, \mathbf{z}^{\prime}, y^{\prime}\right)\right) \sim \gamma}\left[c\left((\mathbf{x}, \mathbf{z}, y),\left(\mathbf{x}^{\prime}, \mathbf{z}^{\prime}, y^{\prime}\right)\right)\right]
\label{eq:wd}
\end{equation}
where $c: \mathcal{X} \times \mathcal{X} \rightarrow[0,+\infty)$ is the convex, lower semicontinuous transport cost function with $c(\mathbf{t}, \mathbf{t})=0$, and $\Pi\left(\hat{P}, P^{*}\right)$ is the set of all distributions whose marginals are given by $\hat{P}$ and $P^{*}$.
\end{definition}
The Wasserstein distance intuitively refers to the minimum cost associated with transporting mass between probability measures.
\begin{prop}
Suppose that the transportation cost $c$ in~\eqref{eq:wd} is continuous and the probability of fair representation is bounded away from zero, i.e., $f_{\phi_1}(\boldsymbol{Z})$, then the minimax objective~\eqref{eq:obj} has a desirable formulation as
\begin{equation}
\small
    \underset{f_{\theta} \in \mathcal{F}}{\operatorname{min}} \sup _{\hat{q}} \mathbb{E}_{P}\left[\frac{\delta\left(Y, f_{\phi_1}(\boldsymbol{Z})\right)}{\hat{q}(\mathbf{Z}|\mathbf{X})}\right]-\lambda W_{c}\left(\hat{q}(\boldsymbol{Z}\mid \boldsymbol{X}), q_{*}\right)
    \label{eq:obj23}
\end{equation}
\label{prop51}
\end{prop}
To make sense of~\eqref{eq:obj23}, we see that while $\hat{q}(\boldsymbol{Z}\mid \boldsymbol{X})$ is acting adversarially against $f_{\phi_1}$ as the inverse weights in the first term, it cannot arbitrarily increase the objective function, since the second terms act as a regularizer that keeps $\hat{q}(\boldsymbol{Z}\mid \boldsymbol{X})$ close to the fair representation $\boldsymbol{Z}$.
The objective loss in~\eqref{eq:obj23} can be converted to a two-model adversarial game:
\begin{equation}
\small 
\underset{f_{\phi_1} \in \mathcal{F}}{\operatorname{min}} \sup _{f_{\phi_2} \in \mathcal{G}} \mathbb{E}_{P}\left[\frac{\mathcal{L}\left(Y, f_{\phi_1}( \boldsymbol{Z})\right)}{G\left(f_{\phi_1}( \boldsymbol{Z})\right)}\right]-\lambda W_{c}\left(G\left(f_{\phi_1}(\mathbf{Z})\right), G\left(f^{*}_{\phi_1}\right)\right)
\label{eq:obj1}
\end{equation}

\begin{theorem}[McDiarmid Inequality]~\cite{mcdiarmid1989method} Let \(\Omega_{1}, \ldots, \Omega_{m}\) be probability spaces. Let \(\Omega=\prod_{k=1}^{m} \Omega_{k}\)
and let \(X\) be a random variable on \(\Omega\) which is uniformly difference-bounded by \(\frac{\lambda}{m} \cdot\) Let
\(\mu=\mathrm{E}(X) .\) Then, for any \(\tau>0\)
\begin{equation}
    P(X-\mu \geq \tau) \leq \exp \left(-\frac{2 \tau^{2} m}{\lambda^{2}}\right)
\end{equation}
\label{eq:mcd}
\end{theorem}
\begin{prop}
Suppose that the transportation cost $c$ is continuous and the probability of fair representation is bounded away from zero, i.e., $\hat{q}(\mathbf{Z}\mid \mathbf{X} )$, then the minimax objective has a desirable formulation as
\begin{equation}
    \underset{f_{\phi_1} \in \mathcal{F}}{\operatorname{min}} \sup _{\hat{q}} \mathbb{E}_{P}\left[\frac{\delta\left(Y, f_{\phi_1}(\mathbf{Z})\right)}{\hat{q}(\mathbf{Z}\mid \mathbf{X} )}\right]-\lambda W_{c}\left(\hat{q}(\mathbf{Z}\mid \mathbf{X}), q_{*}\right)
    \label{eq:obj2}
\end{equation}
\label{prop:2}
\end{prop}
The following theorem discusses the theoretical guarantees for the generalization error of Eq.~\eqref{eq:obj1}.

\begin{theorem}
Suppose the mapping \(G\) from \(f_{\phi_1}\) to \(\hat{q}(\boldsymbol{Z}\mid \boldsymbol{X})\) is one-to-one and surjective with
\(g_{\psi} \in \mathcal{G} .\) Let \(\tilde{\mathcal{G}}(\rho)=\left\{g_{\psi} \in \mathcal{G} \mid W_{c}\left(G\left(g_{\psi}\right), G\left(g^{*}\right)\right) \leq \rho\right\} .\) Then under the conditions specified in Proposition~\ref{prop51}. for all \(\gamma \geq 0\) and \(\rho>0\), the following inequality holds with probability at least \(1-\epsilon\):
\begin{equation}
\small
\begin{split}
\sup _{g_{\psi} \in \tilde{\mathcal{G}}(\rho)} \mathbb{E}_{P}\left[\frac{\mathcal{L}\left(Y, f_{\phi_1}(\boldsymbol{Z})\right)}{G\left(f_{\phi_1}(\boldsymbol{Z})\right)}\right] \leq c_{1} \gamma \rho+\mathbb{E}_{P_{n}}\left[\Delta_{\gamma}\left(f_{\phi_1}; (\boldsymbol{Z}, Y)\right)\right]+\frac{24 \mathcal{J}(\tilde{\mathcal{F}})+c_{2}\left(M, \sqrt{\log \frac{2}{\epsilon}}, \gamma\right)}{\sqrt{n}}
\end{split}
\end{equation}
where $\mathbb{E}_{P_{n}}\left[\Delta_{\gamma}\left(f_{\phi_1}; (\boldsymbol{Z}, Y)\right)\right]$ is a cost-regulated loss given in Proof part below, \(c_{1}\) is a positive constants and \(c_{2}\) is a simple linear function with positive weights. 
\end{theorem}

The above theorem states our main theoretical result on the worst-case generalization bound under the minimax setting.

\begin{proof}
We introduce a cost-regulated loss which is defined as 
\begin{equation}
    \Delta_{\gamma}\left(f_{\phi_1} ;(\mathbf{z}, y)\right)=\sup _{\left( \mathbf{z}^{\prime}, y^{\prime}\right) \in \mathcal{X}}\left\{\frac{\delta\left(y^{\prime},f_{\phi_1}(\mathbf{Z}^{\prime})\right)}{q\left(o=1 \mid  \mathbf{z}^{\prime}\right)}-\right.\left.\gamma c\left((\mathbf{z}, y),\left( \mathbf{z}^{\prime}, y^{\prime}\right)\right)\right\}
\end{equation}
Based on definition of $\Delta_{\gamma}$, we have
\begin{equation}
    \begin{split}
        &\sup _{f_{\phi_1} \in \tilde{\mathcal{G}}(\rho)} \mathbb{E}_{P}\left[\frac{\delta\left(Y, f_{\phi_1}(\mathbf{Z})\right)}{G\left(f_{\phi_1}(\mathbf{X}, \mathbf{Z})\right)}\right]\\
&\leq \inf _{\gamma \geq 0}\left\{\gamma \rho+\int \sup _{\mathbf{h} \in \mathcal{X}}\left(\frac{\delta_{f_{\phi_1}}(\mathbf{h})}{\hat{q}(\mathbf{h})}-\gamma c\left(\mathbf{h}, \mathbf{h}^{\prime}\right)\right) d P(\mathbf{h})\right\}\\
&=\inf _{\gamma \geq 0}\left\{\gamma \rho+\mathbb{E}_{P}\left[\Delta_{\gamma}\left(f_{\phi_1} ; \mathbf{H}\right)\right]\right\} \quad\left(\right. \text{by the definition of} \left.\Delta_{\gamma}\right)\\
&\leq \inf _{\gamma \geq 0}\left\{\gamma \rho+\mathbb{E}_{P_{n}}\left[\Delta_{\gamma}\left(f_{\phi_1} ; \mathbf{H}\right)\right]+\sup _{f_{\phi_1} \in \mathcal{F}}\left(\mathbb{E}_{P}\left[\Delta_{\gamma}\left(f_{\phi_1};\mathbf{H}\right)\right]-\mathbb{E}_{P_{n}}\left[\Delta_{\gamma}\left(f_{\phi_1} ; \mathbf{H}\right)\right]\right)\right\}
    \end{split}
    \label{eq:3}
\end{equation}
Let $W_{\gamma}=\sup _{f_{\phi_1} \in \mathcal{F}}\left(\mathbb{E}_{P}\left[\Delta_{\gamma}\left(f_{\phi_1} ; \mathbf{H}\right)\right]-\mathbb{E}_{P_{n}}\left[\Delta_{\gamma}\left(f_{\phi_1} ; \mathbf{H}\right)\right]\right)$, then we have 
\begin{equation}
    W_{\gamma}=\frac{1}{n} \sup _{f_{\phi_1} \in \mathcal{F}}\left[\sum_{i=1}^{N} \mathbb{E}_{P}\left[\Delta_{\gamma}\left(f_{\phi_1} ; \mathbf{H}\right)\right]-\Delta_{\gamma}\left(f_{\phi_1} ; \mathbf{H}_{i}\right)\right] \quad \gamma \geq 0
    \label{eq:wr}
\end{equation}
According to Theorem~\ref{eq:mcd} and the fact that \(\left|\delta_{f_{\phi_1}}(\mathbf{h})\right| \leq \mu M\) holds uniformly, we have
\begin{equation}
    p\left(W_{\gamma}-\mathbb{E} W_{\gamma} \geq \mu M \sqrt{\frac{\log 1 / \epsilon}{2 N}}\right) \leq \epsilon
        \label{eq:5}
\end{equation}
where \(\epsilon_{1}, \ldots, \epsilon_{N}\) is denoted as the i.i.d Rademacher random variables independent of \(\mathbf{H}\), and \(\mathbf{H}_{i}^{\prime}\) is the i.i.d
copy of \(\mathbf{H}_{i}\) for \(i=1, \ldots, N\).

Considering Eq.~\eqref{eq:wr}, we use the symmetrization argument to reformulate $\mathbb{E} W_{\gamma}$ in Eq.~\eqref{eq:5} as

\begin{equation}
    \begin{aligned} \mathbb{E} W_{\gamma} &=\mathbb{E}\left[\sup _{f_{\phi_1} \in \mathcal{F}}\left|\sum_{i=1}^{N} \Delta_{\gamma}\left(f_{\phi_1} ; \mathbf{H}_{i}^{\prime}\right)-\sum_{i=1}^{N} \Delta_{\gamma}\left(f_{\phi_1} ; \mathbf{H}_{i}\right)\right|\right] \\ &=\mathbb{E}\left[\sup _{f_{\phi_1} \in \mathcal{F}}\left|\frac{1}{N} \sum_{i=1}^{N} \epsilon_{i} \Delta_{\gamma}\left(f_{\phi_1} ; \mathbf{H}_{i}^{\prime}\right)-\frac{1}{N} \sum_{i=1}^{N} \Delta_{\gamma}\left(f_{\phi_1} ; \mathbf{H}_{i}\right)\right|\right] \\ & \leq 2 \mathbb{E}\left[\sup _{f_{\phi_1} \in \mathcal{F}}\left|\frac{1}{N} \sum_{i=1}^{N} \epsilon_{i} \Delta_{\gamma}\left(f_{\phi_1} ; \mathbf{H}_{i}\right)\right|\right] \end{aligned}
\end{equation}
Apparently, each \(\epsilon_{i} \Delta_{\gamma}\left(f_{\phi_1} ; \mathbf{H}_{i}\right)\) is zero-mean, and now we show that it is sub-Gaussian as well.
The bounded difference between two \(f_{\phi_1}, f_{\phi_1}^{\prime}\) is
\begin{equation}
    \begin{split}
        &\mathbb{E}\left[\exp \left(\lambda\left(\frac{1}{\sqrt{N}} \epsilon_{i} \Delta_{\gamma}\left(f_{\phi_1} ; \mathbf{H}_{i}\right)-\frac{1}{\sqrt{N}} \epsilon_{i} \Delta_{\gamma}\left(f_{\phi_1}^{\prime} ; \mathbf{H}_{i}\right)\right)\right)\right]\\
        &=\left(\mathbb{E}\left[\exp \left(\frac{\lambda}{\sqrt{N}} \epsilon_{1}\left(\Delta_{\gamma}\left(f_{\phi_1} ; \mathbf{H}_{1}\right)-\Delta_{\gamma}\left(f_{\phi_1}^{\prime} ; \mathbf{H}_{1}\right)\right)\right)\right]\right)^{N}\\
        &=\left(\mathbb{E}\left[\exp \left(\frac{\lambda}{\sqrt{N}} \epsilon_{1}\left(\sup _{\mathbf{h}^{\prime}} \inf _{\mathbf{h}^{\prime \prime}}\left\{\frac{\delta_{f_{\phi_1}}\left(\mathbf{h}^{\prime}\right)}{q\left(\mathbf{h}^{\prime}\right)}-\gamma c\left(\mathbf{H}_{1}, \mathbf{h}^{\prime}\right)-\frac{\delta_{f_{\phi_1}^{\prime}}\left(\mathbf{h}^{\prime \prime}\right)}{q\left(\mathbf{h}^{\prime \prime}\right)}\right\}+\gamma c\left(\mathbf{H}_{1}, \mathbf{h}^{\prime \prime}\right)\right)\right)\right]\right)^{N}\\
        &\leq\left(\mathbb{E}\left[\exp \left(\frac{\lambda}{\sqrt{N}} \epsilon_{1}\left(\sup _{\mathbf{h}^{\prime}}\left\{\frac{\delta_{f_{\phi_1}}\left(\mathbf{h}^{\prime}\right)}{q\left(\mathbf{h}^{\prime}\right)}-\frac{\delta_{f_{\phi_1}^{\prime}}\left(\mathbf{h}^{\prime}\right)}{q\left(\mathbf{h}^{\prime}\right)}\right\}\right)\right)\right]\right)^{N}\\
        &\leq \exp \left(\lambda^{2}\left\|\frac{\delta_{f_{\phi_1}}}{q}-\frac{\delta_{f_{\phi_1}^{\prime}}}{q}\right\|_{\infty}^{2} / 2\right) \quad \text{(by Hoeffding's inequality)}
    \end{split}
\end{equation}
Hence we see that \(\frac{1}{\sqrt{N}} \epsilon_{i} \Delta_{\gamma}\left(f_{\phi_1} ; \mathbf{H}_{i}\right)\) is sub-Gaussian with respect to \(\left\|\frac{\delta_{f_{\phi_1}}}{q}-\frac{\delta_{f_{\phi_1}^{\prime}}}{q}\right\|_{\infty}^{2} .\) Therefore,
\(\mathbb{E} W_{\gamma}\) can be bounded| by using the standard technique for Rademacher complexity and Dudley's
entropy integral
\begin{equation}
    \mathbb{E} W_{\gamma} \leq \frac{24}{N} \mathcal{J}(\tilde{\mathcal{F}})
        \label{eq:8}
\end{equation}
Based on all above bounds in ~\eqref{eq:3}, ~\eqref{eq:5} and ~\eqref{eq:8} we obtain the desired result.
\end{proof}

\section{Experiments}\label{sec:expr}

Compared to other fairness criteria, evaluating the performance of counterfactual fairness is frustratingly difficult due to the absence of ground truth samples. In fact, from the observational data, we are unable to observe the characteristic of individuals in the counterfactual world where we make an intervention into their sensitive attributes. In fact, we cannot simply change the values of sensitive attributes since the intervention on the sensitive features can lead to changes in some non-sensitive features due to the causal effects. For example, we have an observational individual $x$, but do not have its counterfactual version $\hat{x}$; therefore, it is not feasible to evaluate the performance of predictive model $f(.)$ by measuring the similarity of $f(x)$ and $f(\hat{x})$. In the previous studies \cite{kusner2017counterfactual,russell2017worlds,wu2019counterfactual}, they generate both the original samples and counterfactual samples from the structural causal model. However, it is hard to verify the trustworthiness of the samples due to the unidentifiability of the causal model. In this research, by getting a pair of similar individuals sharing the same properties, we thus can approximately evaluate the model performance. This means that instead of evaluating in the counterfactual space, we can approximately evaluate the performance of counterfactual fairness via the individual fairness criteria. We conducted extensive experiments on three real-world datasets with different evaluation metrics for two tasks including regression and classification tasks.

\subsection{Datasets}


We evaluate our approach via regression datasets including \texttt{LSAC} \cite{wightman1998lsac} and classification datasets including \texttt{Compas} \cite{larson2016we} and \texttt{Adult}.

\begin{itemize}
    \item \texttt{LSAC}\footnote{Download at: \url{http://www.seaphe.org/databases.php}} \cite{wightman1998lsac}. \texttt{LSAC} dataset provides information about law students including their gender, race, entrance exam scores (LSAT), grade-point average (GPA) and first-year average grade (FYA). The main task is to determine which applicants would have a high possibility to obtain high FYA. The school also ensures that model decisions are not biased by sensitive attribtues including race and gender. We pay attention to predict the FYA of a student. 
    \item \texttt{Compas}\footnote{Download at: \url{https://www.propublica.org}} \cite{larson2016we}. \texttt{Compas} dataset has been released by ProPublica about prisoners in Florida (US) and also has been previously explored for fairness studies in criminal justice \cite{berk2021fairness}. The dataset contains information about 6,167 prisoners, and each individual has two sensitive attributes including gender, race and other attributes related to prior conviction and age. The main task is to predict whether or not a prisoner will re-offend within two years after being released from prison.  
    \item \texttt{Adult}\footnote{Download at: \url{https://archive.ics.uci.edu/ml/datasets/adult}}\cite{Dua:2019}. \texttt{Adult} dataset is the real-world dataset providing information about loan applicants in the financial organization. The dataset consists of both continuous features and categorical features. The main task is to determine whether a person has an annual income exceeding \$50k dollars. The sensitive attributes are gender and race. 
\end{itemize}

To evaluate the generalization capability of models, we randomly split each dataset into 80\% training and 20\% test set. We conduct 100 repeated experiments, then evaluate performance on the test set and finally report the average statistics. 

\begin{figure}[!htb]
\minipage{0.32\textwidth}
  \includegraphics[width=\linewidth]{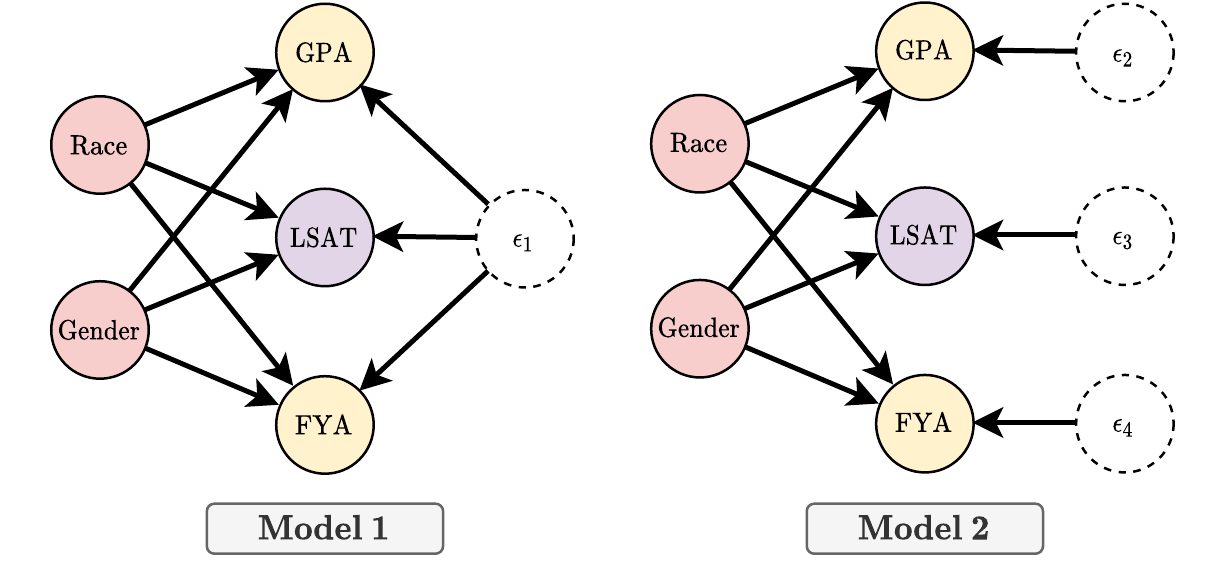}
\caption*{\small \texttt{Law} dataset}
\endminipage\hfill
\minipage{0.32\textwidth}
  \includegraphics[width=\linewidth]{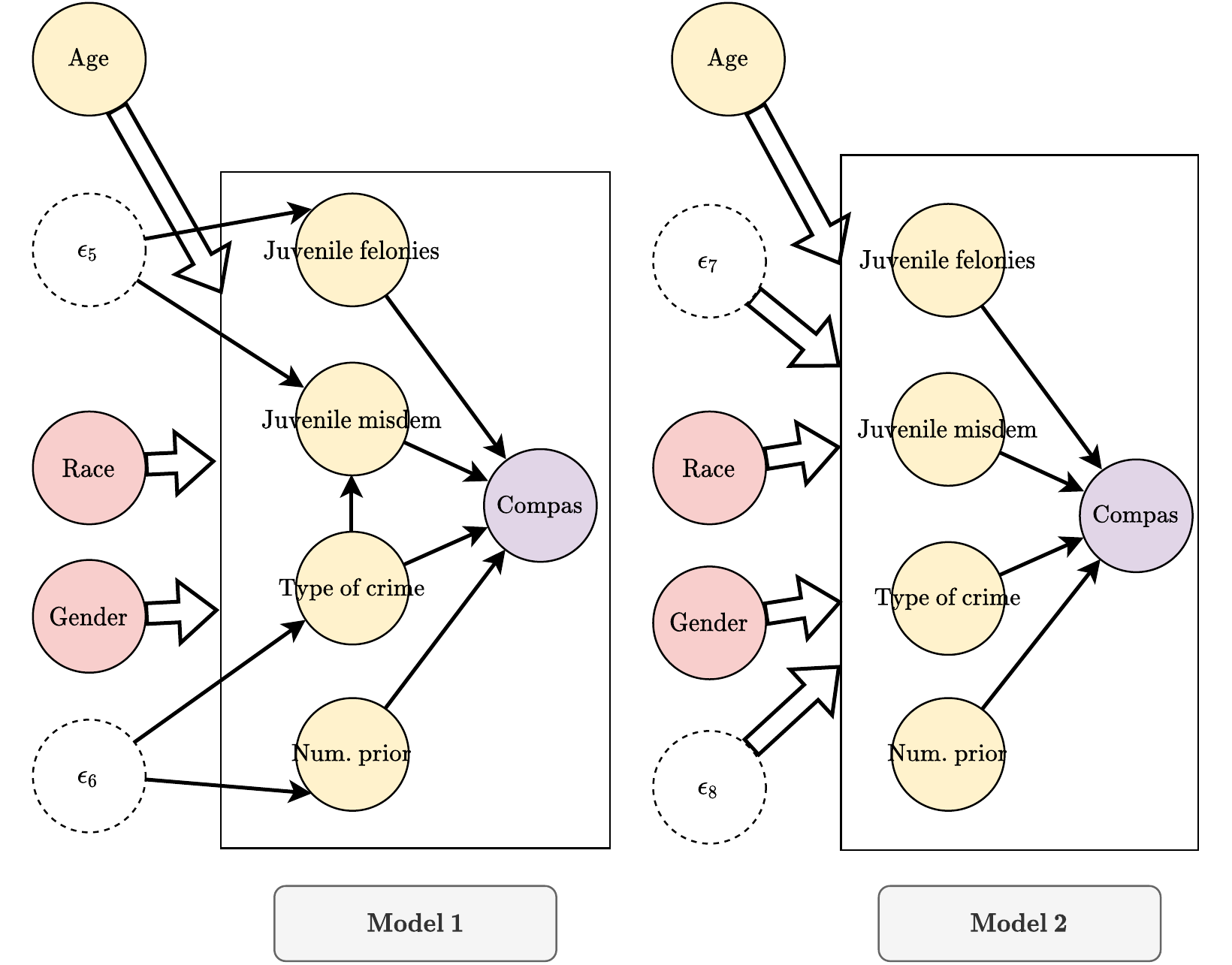}
  \caption*{\small Compas dataset}
\endminipage\hfill
\minipage{0.32\textwidth}%
  \includegraphics[width=\linewidth]{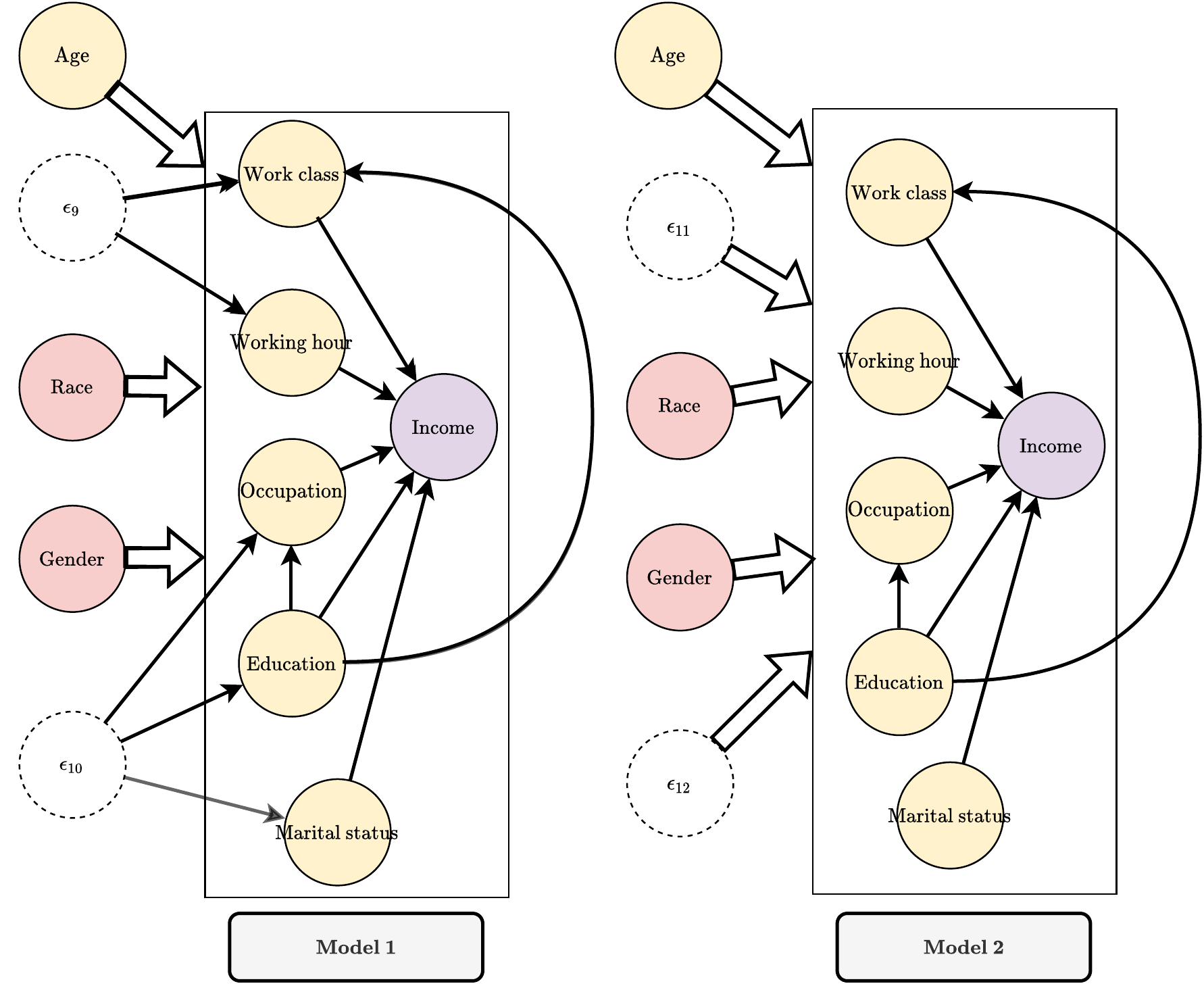}
\caption*{\small \texttt{Adult} dataset}
\endminipage
\caption{Causal diagrams for Law, Compas and \texttt{Adult} dataset. $\{\epsilon_1\cdots \epsilon_{12}\}$ are the unobserved variables. The large white arrows represent that each variable has a causal effect on every variables contained in the box.}
\label{fig:scm}
\end{figure}


\subsection{Baselines}
We make a comparison with several state-to-the-art methods as below.

\begin{itemize}
    \item \textbf{Full features (Full)} \cite{kusner2017counterfactual} is the standard technique that uses all the features including both the sensitive and non-sensitive ones.
    \item \textbf{Unaware features (Unaware)} \cite{chen2019fairness} does not consider sensitive features such as race or gender in the input and only utilizes non-sensitive features.
    \item \textbf{Counterfactual fairness model (CF)} \cite{kusner2017counterfactual} uses the causal graph and infers the latent variables which are not the child nodes of the sensitive features. Since there is no ground truth causal model, we consider two causal diagrams illustrated in Figure~\ref{fig:scm} for each dataset as \textbf{CF$_1$} and \textbf{CF$_2$}. \textbf{CF$_1$} and \textbf{CF$_2$} correspond to two different structural causal models
    \item \textbf{Multi-world models (Multi-wolrd)} \cite{russell2017worlds} minimizes the model predictions when considering different structural causal models. Specifically, with two causal diagrams in Figure~\ref{fig:scm} for each dataset, we build a neural network and use the gradient descent algorithm to minimize the model output from two causal models. 
    \item \textbf{Auto-encoder model (AE)} \cite{ng2011sparse} uses the encoder-decoder architecture to learn the latent representation. This model utilizes all the features including the sensitive and non-sensitive ones as the input. 
\end{itemize}
Note that \textbf{Full}, \textbf{Unaware}, \textbf{CF$_1$}, \textbf{CF$_2$} and \textbf{Autoencoder} are the representation methods that only produce features, so we use these features as an input and construct predictive models including Linear Regression \textbf{(LR)} and Gradient Boosting Regression \textbf{(GBboostR)} for regression task, and Logistic Regression \textbf{(Log)} and Gradient Boosting Classifier \textbf{(GBboostC)} for the classification task. As our method aims to output a fair and informative representation, we use two models: Invariant-encoder model and Fair-learning predictive model. In order to gain more insights into model behaviors, we first utilize {Invariant-encoder model} \textbf{(InvEnc)} to generate the latent representation, and also combine with {LR}, {GBboostR}, and {GBboostC}. Finally, we adopt the fair-learning predictive model and invariant-encoder together, referred as \textbf{InvFair}. 

\begin{table*}[t]
\begin{adjustbox}{width=1.\columnwidth,center}
\begin{tabular}{@{}cccccc@{}}
\toprule
\multirow{2}{*}{\textbf{Method}} & \multicolumn{3}{c}{\textbf{Regression metric}}     & \multicolumn{2}{c}{\textbf{Fairness metric}}       \\ \cmidrule(lr){2-4} \cmidrule(lr){5-6} 
                                 & \textbf{RMSE}                & \textbf{MAE}                 & \textbf{R2score}              & \textbf{Wasserstein}            & \textbf{Gaussian}            \\ \midrule
Full-LR                         & \textbf{0.870 $\pm$ 3.2e-03} & \textbf{0.705 $\pm$ 7.4e-03} & 0.120 $\pm$ 5.0e-03           & 0.522 $\pm$ 2.2e-03          & 0.719 $\pm$ 3.2e-03          \\
Full-GBoostR                     & 0.935 $\pm$ 2.8e-03          & 0.751 $\pm$ 3.2e-03          & \textbf{-0.014 $\pm$ 4.9e-03} & 0.010 $\pm$ 6.5e-03          & 0.037 $\pm$ 5.5e-03          \\ \midrule
Unaware-LR                      & 0.889 $\pm$ 7.5e-03          & 0.718 $\pm$ 3.2e-03          & 0.083 $\pm$ 2.4e-03           & 0.097 $\pm$ 4.5e-03          & 0.194 $\pm$ 2.1e-03          \\
Unaware-GBoostR                  & 1.034 $\pm$ 7.8e-03          & 0.829 $\pm$ 5.1e-03          & \textbf{-0.242 $\pm$ 3.2e-03} & \textbf{0.009 $\pm$ 4.8e-03}          & 0.030 $\pm$ 5.1e-03          \\ \midrule
CF$_1$-LR                          & 0.906 $\pm$ 4.1e-03          & 0.730 $\pm$ 5.1e-03          & 0.048 $\pm$ 4.8e-03           & 0.019 $\pm$ 3.1e-03          & 0.045 $\pm$ 3.0e-03          \\
CF$_1$-GBoostR                     & 0.909 $\pm$ 2.1e-03          & 0.732 $\pm$ 4.6e-03          & 0.0410 $\pm$ 5.1e-03           & 0.013 $\pm$ 7.6e-03          & 0.037 $\pm$ 4.3e-03          \\
CF$_2$-LR                       & 0.914 $\pm$ 7.3e-03          & 0.736 $\pm$ 5.1e-03          & 0.030 $\pm$ 6.4e-03           & 0.070 $\pm$ 7.5e-03          & 0.030 $\pm$ 8.1e-03          \\
CF$_2$-GBoostR                      & 0.913 $\pm$ 4.9e-03          & 0.734 $\pm$ 3.6e-03          & 0.034 $\pm$ 7.5e-03           & 0.070 $\pm$ 7.3e-03          & 0.032 $\pm$ 8.5e-03          \\
Multi-world                      & 0.917 $\pm$ 3.9e-03          & 0.736 $\pm$ 7.1e-03          & 0.025 $\pm$ 7.1e-03           & 0.030 $\pm$ 5.8e-03          & 0.036 $\pm$ 4.7e-03          \\
\midrule
AE-LR                           & \textbf{0.870 $\pm$ 7.1e-03} & \textbf{0.705 $\pm$ 4.1e-03} & 0.121 $\pm$ 6.0e-03           & 0.532 $\pm$ 2.1e-03          & 0.705 $\pm$ 8.1e-03          \\
AE-GBoostR                       & 0.889 $\pm$ 3.6e-03          & 0.715 $\pm$ 8.1e-03          & 0.221 $\pm$ 5.1e-03           & 0.425 $\pm$ 8.1e-03          & 0.815 $\pm$ 4.8e-03          \\
InvEnc-LR            & 0.905 $\pm$ 3.4e-03          & 0.727 $\pm$ 7.1e-03          & 0.040 $\pm$ 2.1e-03           & 0.131 $\pm$ 8.1e-03          & 0.160 $\pm$ 5.4e-03          \\
InvEnc-GBoostR        & 0.904 $\pm$ 7.1e-03          & 0.773 $\pm$ 3.1e-03          & 0.131 $\pm$ 3.2e-03           & 0.183 $\pm$ 1.9e-03          & 0.179 $\pm$ 7.1e-03          \\
\midrule
InvFair (Ours)      & 0.900 $\pm$ 2.2e-03          & 0.739 $\pm$ 2.5e-03          & 0.087 $\pm$ 4.1e-03           & \textbf{0.009 $\pm$ 8.6e-03} & \textbf{0.029 $\pm$ 2.1e-03} \\ \bottomrule
\end{tabular}
\end{adjustbox}
\caption{Performance comparisons on \texttt{Law} dataset. The mean and variance for each method are obtained via 100 repeated runs. For \textbf{R2score}, results in bold font show the corresponding models are unreliable. For the \textbf{remaining metrics}, the best results are bold. For each method, we use (baseline)-LR/GBoostR to show the baseline combined with Logistic regression or Gradient boosting.}
\label{table:law}
\end{table*}

\begin{table*}[t]
\begin{adjustbox}{width=0.7\columnwidth,center}
\begin{tabular}{@{}cccccc@{}}
\toprule
\multirow{2}{*}{\textbf{Method}} & \multicolumn{3}{c}{\textbf{Regression metric}}     & \multicolumn{2}{c}{\textbf{Fairness metric}}       \\ \cmidrule(lr){2-4} \cmidrule(lr){5-6} 
                                 & \textbf{RMSE}                & \textbf{MAE}                 & \textbf{R2score}              & \textbf{Wasserstein}            & \textbf{Gaussian}            \\ \midrule
Full-LR                          & 0.0356                                             & 0.0178                                            & 0.0227                                                & 0.0388                                                    & 0.019                                                  \\
Full-GBoostR                     & 0.0262                                             & 0.0475                                            & 0.031                                                 & 0.0433                                                    & 0.034                                                  \\ \midrule
Unaware-LR                       & 0.0106                                             & 0.0085                                            & 0.0234                                                & 0.0161                                                    & 0.0432                                                 \\
Unaware-GBoostR                  & 0.0371                                             & 0.0324                                            & 0.0322                                                & 0.028                                                     & 0.0392                                                 \\ \midrule
CF$_1$-LR                           & 0.0416                                             & 0.0302                                            & 0.0251                                                & 0.0262                                                    & 0.0495                                                 \\
CF$_1$-GBoostR                      & 0.0115                                             & 0.0214                                            & 0.0089                                                & 0.0161                                                    & 0.0215                                                 \\
CF$_2$-LR                           & 0.0449                                             & 0.0325                                            & 0.041                                                 & 0.0253                                                    & 0.0376                                                 \\
CF$_2$-GBoostR                      & 0.0444                                             & 0.0308                                            & 0.0314                                                & 0.0464                                                    & 0.0479                                                 \\
Multi-world                       & 0.0253                                             & 0.0377                                            & 0.0440                                                 & 0.0402                                                    & 0.0136                                                 \\ \midrule
AE-LR                            & 0.0426                                             & 0.0124                                            & 0.0254                                                & 0.0229                                                    & 0.0284                                                 \\
AE-GBoostR                       & 0.0438                                             & 0.0238                                            & 0.0264                                                & 0.0251                                                    & 0.0289                                                 \\
InvEnc-LR                        & 0.0162                                             & 0.0321                                            & 0.0215                                                & 0.0154                                                    & 0.0278                                                 \\
InvEnc-GBoostR                   & 0.0285                                             & 0.0474                                            & 0.0259                                                & 0.0489                                                    & 0.0349                                         \\         \bottomrule                                          
\end{tabular}
\end{adjustbox}
\caption{We compute $p$-value by conducting a paired $t$-test between our approach and baselines with 100 repeated experiments for each metric on Law dataset.}
\label{tab:p1}
\end{table*}

\subsection{Evaluation metrics}
Our method aims to learn the fair and informative representation that can be used for downstream classification or regression.
We use two metrics for prediction and fairness performance, and consider both regression and classification tasks.

For the prediction performance, we use root mean squared error {(RMSE)} and mean absolute error {(MAE)} for the {regression task}. 
For the {classification task}, we use {Precision}, {Recall}, {F$_1$ score}, and {Balanced Accuracy} \cite{brodersen2010balanced} for evaluation purpose. 
We emphasize that since the \texttt{Adult} and \texttt{Compas} datasets are highly imbalanced, we use the Balanced Accuracy instead of the traditional accuracy, which is defined as $\text{Balanced Acc} = \frac{\text{TPR} + \text{TNR}}{2}$ where \text{TPR} and \text{TNR} are true positive rate, and true negative rate, respectively.

For the fairness performance, we use Wasserstein distance (Wasserstein) \cite{ruschendorf1985wasserstein} and maximum mean discrepancy (MMD) with Gaussian kernel \cite{gretton2012kernel,oh2019kernel} (Gaussian) in the regression task. On the other hand, we utilize generalized entropy index \cite{speicher2018unified} to evaluate the performance in the classification task. Generalized entropy index that has been previously used widely in economics is explored by \cite{speicher2018unified} as the unified approach to evaluate the performance of individual fairness algorithms defined as follows:




     


\begin{equation} \small
    \begin{split}\mathcal{E}(\alpha) = \begin{cases}
    \frac{1}{n \alpha (\alpha-1)}\sum_{i=1}^n\left[\left(\frac{b_i}{\mu}\right)^\alpha - 1\right],& \alpha \ne 0, 1,\\
    \frac{1}{n}\sum_{i=1}^n\frac{b_{i}}{\mu}\ln\frac{b_{i}}{\mu},& \alpha=1,\\
    -\frac{1}{n}\sum_{i=1}^n\ln\frac{b_{i}}{\mu},& \alpha=0.
\end{cases}\end{split}
\end{equation}
where $b_i = \hat{y}_i - y_i + 1
$, $\mu = \frac{1}{n}\sum_i^n b_i$. In this study, we use  $\mathcal{E}(1)$ and $\mathcal{E}(2)$ which are called Theil index {(TI)} and coefficient of variation {(CV)}, respectively.
For all metrics except Precision, Recall, F$_1$ score, and Balanced Accuracy, lower values are better.

\begin{table*}[!htb]
\begin{adjustbox}{width=1.\columnwidth,center}
\begin{tabular}{@{}ccccccc@{}}
\toprule
\multirow{2}{*}{\textbf{Method}} & \multicolumn{4}{c}{\textbf{Classification metric}}                                             & \multicolumn{2}{c}{\textbf{Fairness metric}}    
\\
        \cmidrule(lr){2-5} \cmidrule(lr){6-7} 
                                 & \textbf{Balanced Acc}                & \textbf{F$_1$}                 & \textbf{Precision}              & \textbf{Recall}            & \textbf{CV}            & \textbf{TI}           \\
\midrule
Full-Log    & 0.660 $\pm$ 5.3e-03                           & 0.664 $\pm$ 7.5e-03                           & 0.672 $\pm$ 3.1e-03                           & 0.670 $\pm$ 3.0e-03                           & 0.891 $\pm$ 4.5e-03                           & 0.285 $\pm$ 2.8e-03                           \\
Full-GBoostC      & 0.665 $\pm$ 1.8e-03                           & 0.670 $\pm$ 7.5e-03                           & 0.675 $\pm$ 7.8e-03                           & 0.674 $\pm$ 7.6e-03                           & 0.872 $\pm$ 6.5e-03                           & 0.271 $\pm$ 8.2e-03                           \\ \midrule
Unaware-Log & 0.662 $\pm$ 4.5e-03                           & 0.666 $\pm$ 7.9e-03                           & 0.674 $\pm$ 7.9e-03                           & 0.672 $\pm$ 6.0e-03                           & 0.887 $\pm$ 1.8e-03                           & 0.282 $\pm$ 4.6e-03                           \\
Unaware-GBoostC   & 0.662 $\pm$ 7.9e-03                           & 0.666 $\pm$ 7.7e-03                           & 0.672 $\pm$ 4.2e-03                           & 0.672 $\pm$ 4.9e-03                           & 0.880 $\pm$ 3.1e-03                           & 0.276 $\pm$ 5.0e-03                           \\  \midrule
CF$_1$-Log     & 0.500 $\pm$ 8.0e-03                           & 0.381 $\pm$ 2.3e-03                           & 0.522 $\pm$ 5.8e-03                           & 0.540 $\pm$ 4.6e-03                           & 1.306 $\pm$ 6.5e-03                           & 0.615 $\pm$ 1.4e-03                           \\
CF$_1$-GBoost       & 0.534 $\pm$ 3.7e-03                           & 0.517 $\pm$ 2.7e-03                           & 0.551 $\pm$ 5.4e-03                           & 0.556 $\pm$ 1.4e-03                           & 1.159 $\pm$ 3.2e-03                           & 0.463 $\pm$ 6.0e-03                           \\
CF$_2$-Log     & 0.623 $\pm$ 2.5e-03                           & 0.627 $\pm$ 5.7e-03                           & 0.628 $\pm$ 7.5e-03                           & 0.629 $\pm$ 7.2e-03                           & 0.904 $\pm$ 3.7e-03                           & 0.286 $\pm$ 6.8e-03                           \\
CF$_2$-GBoost       & 0.573 $\pm$ 8.5e-03                           & 0.572 $\pm$ 5.6e-03                           & 0.576 $\pm$ 5.5e-03                           & 0.571 $\pm$ 3.4e-03                           & 0.871 $\pm$ 4.7e-03                           & 0.262 $\pm$ 6.9e-03                           \\
Multi-world     & 0.500 $\pm$ 7.5e-03                           & 0.381 $\pm$ 8.6e-03                           & 0.522 $\pm$ 2.1e-03                           & 0.540 $\pm$ 8.5e-03                           & 1.306 $\pm$ 4.5e-03                           & 0.615 $\pm$ 3.6e-03                           \\
\midrule
AE-Log               & 0.659 $\pm$ 6.2e-03                           & 0.663 $\pm$ 4.8e-03                           & 0.667 $\pm$ 4.3e-03                           & 0.667 $\pm$ 5.4e-03                           & 0.876 $\pm$ 6.9e-03                           & 0.272 $\pm$ 4.0e-03                           \\
AE-GBoostC             & 0.666 $\pm$ 5.3e-03                           & 0.670 $\pm$ 6.4e-03                           & 0.676 $\pm$ 6.8e-03                           & 0.675 $\pm$ 8.6e-03                           & 0.874 $\pm$ 5.8e-03                           & 0.273 $\pm$ 1.6e-03                           \\
InvEnc-Log               & \textbf{0.670 $\pm$ 1.5e-03} & \textbf{0.675 $\pm$ 2.9e-03} & \textbf{0.681 $\pm$ 6.8e-03} & \textbf{0.680 $\pm$ 8.5e-03} & 0.869 $\pm$ 4.1e-03                           & 0.270 $\pm$ 1.6e-03                           \\
InvEnc-GBoostC             & 0.666 $\pm$ 5.8e-03                           & 0.670 $\pm$ 1.4e-03                            & 0.676 $\pm$ 8.7e-03                           & 0.675 $\pm$ 5.7e-03                           & 0.874 $\pm$ 6.3e-03                           & 0.273 $\pm$ 2.1e-03                           \\ \midrule
InvFair (Ours)               & 0.668 $\pm$ 2.7e-03                           & 0.672 $\pm$ 2.5e-03                           & 0.672 $\pm$ 2.6e-03                           & 0.673 $\pm$ 5.9e-03                           & \textbf{0.836 $\pm$ 5.1e-03} & \textbf{0.211 $\pm$ 2.2e-03}
\\\bottomrule
\end{tabular}
\end{adjustbox}
\caption{Performance comparison on \texttt{Compas} dataset. The mean and variance for each method are obtained via 100 repeated experiments. The best results are bold. For each method, we name (*)-Log/GBoostC with (*) representing the baseline method.}
\label{table:compas}
\end{table*}

\begin{table*}[!htb]
\label{tab:pvaluecompas}
\begin{adjustbox}{width=0.7\columnwidth,center}
\begin{tabular}{@{}ccccccc@{}}
\toprule
\multirow{2}{*}{\textbf{Method}} & \multicolumn{4}{c}{\textbf{Classification metric}}                                             & \multicolumn{2}{c}{\textbf{Fairness metric}}    
\\
        \cmidrule(lr){2-5} \cmidrule(lr){6-7} 
                                 & \textbf{Balanced Acc}                & \textbf{F$_1$}                 & \textbf{Precision}              & \textbf{Recall}            & \textbf{CV}            & \textbf{TI}           \\
\midrule
Full-Log                         & 0.0419                                                     & 0.0498                                           & 0.040                                                    & 0.0489                                               & 0.0214                                           & 0.031                                            \\
Full-GBoostC                     & 0.0112                                                     & 0.0482                                           & 0.0101                                                  & 0.0261                                               & 0.014                                            & 0.027                                            \\ \midrule
Unaware-Log                      & 0.0106                                                     & 0.0450                                            & 0.0458                                                  & 0.0342                                               & 0.047                                            & 0.0184                                           \\
Unaware-GBoostC                  & 0.0476                                                     & 0.0164                                           & 0.0347                                                  & 0.0364                                               & 0.0391                                           & 0.0159                                           \\ \midrule
CF$_1$-Log                          & 0.0364                                                     & 0.0367                                           & 0.0323                                                  & 0.017                                                & 0.0305                                           & 0.0173                                           \\
CF$_1$-GBoost                       & 0.0386                                                     & 0.0302                                           & 0.0331                                                  & 0.0144                                               & 0.0105                                           & 0.049                                            \\
CF$_2$-Log                          & 0.0343                                                     & 0.0475                                           & 0.0459                                                  & 0.0122                                               & 0.0384                                           & 0.0148                                           \\
CF$_2$-GBoost                       & 0.0184                                                     & 0.0259                                           & 0.0104                                                  & 0.0173                                               & 0.0475                                           & 0.0302                                           \\
Multi-world                       & 0.0258                                                     & 0.0129                                           & 0.0473                                                  & 0.0186                                               & 0.0316                                           & 0.0344                                           \\ \midrule
AE-Log                           & 0.0164                                                     & 0.0363                                           & 0.0166                                                  & 0.0468                                               & 0.0454                                           & 0.0151                                           \\
AE-GBoostC                       & 0.0401                                                     & 0.0387                                           & 0.0183                                                  & 0.0207                                               & 0.0335                                           & 0.0171                                           \\
InvEnc-Log                       & 0.0208                                                     & 0.036                                            & 0.0272                                                  & 0.0147                                               & 0.0481                                           & 0.0398                                           \\
InvEnc-GBoostC                   & 0.0377                                                     & 0.0364                                           & 0.0157         & 0.030                                                 & 0.0117                                           & 0.0333 
\\\bottomrule
\end{tabular}
\end{adjustbox}
\caption{We compute $p$-value by conducting a paired $t$-test between our approach and baselines with 100 repeated experiments for each metric on Compas dataset.}
\label{tab:p2}
\end{table*}

\begin{table*}[!htb]
\begin{adjustbox}{width=1.\columnwidth,center}
\begin{tabular}{@{}ccccccc@{}}
\toprule
\multirow{2}{*}{\textbf{Method}} & \multicolumn{4}{c}{\textbf{Classification metric}}                                             & \multicolumn{2}{c}{\textbf{Fairness metric}}    
\\
        \cmidrule(lr){2-5} \cmidrule(lr){6-7} 
                                 & \textbf{Balanced Acc}                & \textbf{F$_1$}                 & \textbf{Precision}              & \textbf{Recall}            & \textbf{CV}            & \textbf{TI}           \\
\midrule
Full-Log    & 0.606 $\pm$ 7.6e-03          & 0.741 $\pm$ 6.4e-03                          & 0.743 $\pm$ 6.5e-03          & 0.771 $\pm$ 3.4e-03                           & 0.745 $\pm$ 8.7e-03          & 0.215 $\pm$ 7.7e-03          \\
Full-GBoostC      & 0.730 $\pm$ 6.5e-03           & \textbf{0.820 $\pm$ 1.8e-03} & 0.818 $\pm$ 6.2e-03          & \textbf{0.827 $\pm$ 3.8e-03} & 0.616 $\pm$ 3.4e-03          & 0.142 $\pm$ 3.4e-03          \\  \midrule
Unaware-Log & 0.551 $\pm$ 1.9e-03          & 0.700 $\pm$ 7.8e-03                          & 0.697 $\pm$ 7.2e-03          & 0.747 $\pm$ 2.5e-03                           & 0.801 $\pm$ 8.3e-03          & 0.249 $\pm$ 8.6e-03          \\
Unaware-GBoostC   & 0.725 $\pm$ 7.8e-03          & 0.816 $\pm$ 8.4e-03                          & 0.815 $\pm$ 7.8e-03          & 0.824 $\pm$ 7.2e-03                           & 0.622 $\pm$ 3.7e-03          & 0.145 $\pm$ 6.7e-03          \\ \midrule
CF$_1$-Log     & 0.515 $\pm$ 8.6e-03          & 0.670 $\pm$ 5.8e-03                          & 0.675 $\pm$ 2.3e-03          & 0.749 $\pm$ 8.2e-03                           & 0.814 $\pm$ 3.2e-03          & 0.272 $\pm$ 8.4e-03          \\
CF$_1$-GBR          & 0.513 $\pm$ 2.1e-03          & 0.666 $\pm$ 3.3e-03                          & 0.712 $\pm$ 6.8e-03          & 0.757 $\pm$ 2.4e-03                           & 0.801 $\pm$ 7.0e-03          & 0.273 $\pm$ 6.0e-03          \\
CF$_2$-Log     & 0.515 $\pm$ 4.1e-03          & 0.669 $\pm$ 1.3e-03                          & 0.671 $\pm$ 7.7e-03          & 0.747 $\pm$ 6.4e-03                           & 0.817 $\pm$ 8.0e-03          & 0.272 $\pm$ 3.7e-03          \\
CF$_2$-GBoostC       & 0.520 $\pm$ 5.2e-03          & 0.674 $\pm$ 1.9e-03                          & 0.700 $\pm$ 5.5e-03          & 0.756 $\pm$ 8.1e-03                           & 0.802 $\pm$ 6.4e-03          & 0.269 $\pm$ 3.6e-03          \\
Multi-world     & 0.510 $\pm$ 3.9e-03          & 0.664 $\pm$ 5.1e-03                          & 0.664 $\pm$ 7.0e-03            & 0.747 $\pm$ 2.5e-03                           & 0.819 $\pm$ 3.6e-03          & 0.275 $\pm$ 7.8e-03          \\
\midrule
AE-Log                & 0.730 $\pm$ 2.1e-03          & 0.817 $\pm$ 6.0e-03                          & 0.815 $\pm$ 4.2e-03          & 0.823 $\pm$ 8.2e-03                           & 0.819 $\pm$ 8.2e-03          & 0.242 $\pm$ 2.3e-03          \\
AE-GBoostC             & 0.724 $\pm$ 5.5e-03          & 0.815 $\pm$ 1.9e-03                          & 0.814 $\pm$ 7.6e-03          & 0.823 $\pm$ 2.7e-03                           & 0.823 $\pm$ 3.3e-03          & 0.245 $\pm$ 5.4e-03          \\
InvEnc-Log              & 0.723 $\pm$ 7.9e-03          & 0.812 $\pm$ 6.9e-03                          & 0.810 $\pm$ 4.9e-03           & 0.819 $\pm$ 6.2e-03                           & 0.627 $\pm$ 8.1e-03          & 0.146 $\pm$ 2.8e-03          \\
InvEnc-GBoostC             & 0.724 $\pm$ 3.6e-03          & 0.815 $\pm$ 2.3e-03                          & 0.814 $\pm$ 8.2e-03          & 0.823 $\pm$ 1.4e-03                           & 0.623 $\pm$ 2.6e-03          & 0.145 $\pm$ 7.4e-03          \\  \midrule
InvFair (Ours)                & \textbf{0.778 $\pm$ 8.6e-03} & 0.728 $\pm$ 5.4e-03                          & \textbf{0.835 $\pm$ 3.5e-03} & 0.707 $\pm$ 7.2e-03                           & \textbf{0.556 $\pm$ 4.7e-03} & \textbf{0.090 $\pm$ 7.4e-03} 
\\\bottomrule
\end{tabular}
\end{adjustbox}
\caption{Performance comparison on \texttt{Adult} dataset. The mean and variance for each method are obtained via 100 repeated experiments. The best results are bold. For each method, we name (*)-Log/GBoostC with (*) representing features generated by baseline method.}
\label{table:adult}
\end{table*}

\begin{table*}[!htb]
\begin{adjustbox}{width=0.7\columnwidth,center}
\begin{tabular}{@{}ccccccc@{}}
\toprule
\multirow{2}{*}{\textbf{Method}} & \multicolumn{4}{c}{\textbf{Classification metric}}                                             & \multicolumn{2}{c}{\textbf{Fairness metric}}    
\\
        \cmidrule(lr){2-5} \cmidrule(lr){6-7} 
                                 & \textbf{Balanced Acc}                & \textbf{F$_1$}                 & \textbf{Precision}              & \textbf{Recall}            & \textbf{CV}            & \textbf{TI}           \\
\midrule
Full-Log                         & 0.0288                                                     & 0.0341                                           & 0.0443                                                  & 0.0145                                               & 0.014                                            & 0.0257                                           \\
Full-GBoostC                     & 0.0275                                                     & 0.0149                                           & 0.0458                                                  & 0.0193                                               & 0.0111                                           & 0.0091                                           \\ \midrule
Unaware-Log                      & 0.0122                                                     & 0.0173                                           & 0.0151                                                  & 0.0398                                               & 0.038                                            & 0.0478                                           \\
Unaware-GBoostC                  & 0.0369                                                     & 0.0476                                           & 0.0431                                                  & 0.0385                                               & 0.0177                                           & 0.0352                                           \\ \midrule
CF$_1$-Log                          & 0.0473                                                     & 0.0366                                           & 0.0144                                                  & 0.0222                                               & 0.0446                                           & 0.0337                                           \\
CF$_1$-GBoost                       & 0.0473                                                     & 0.0338                                           & 0.0193                                                  & 0.0492                                               & 0.0109                                           & 0.0298                                           \\
CF$_2$-Log                          & 0.0405                                                     & 0.0269                                           & 0.0393                                                  & 0.0441                                               & 0.0254                                           & 0.0203                                           \\
CF$_2$-GBoost                       & 0.0361                                                     & 0.0348                                           & 0.021                                                   & 0.0151                                               & 0.0107                                           & 0.0233                                           \\
Multi-world                       & 0.0378                                                     & 0.0264                                           & 0.011                                                   & 0.0137                                               & 0.0326                                           & 0.0242                                           \\ \midrule
AE-Log                           & 0.0369                                                     & 0.0307                                           & 0.0325                                                  & 0.0092                                               & 0.0143                                           & 0.013                                            \\
AE-GBoostC                       & 0.0285                                                     & 0.0294                                           & 0.0155                                                  & 0.0409                                               & 0.0175                                           & 0.025                                            \\
InvEnc-Log                       & 0.0372                                                     & 0.0321                                           & 0.048                                                   & 0.0258                                               & 0.0311                                           & 0.0164                                           \\
InvEnc-GBoostC                   & 0.0239                                                     & 0.0466                                           & 0.0218                                                  & 0.0456                                               & 0.0218                                           & 0.0253                                       
\\\bottomrule
\end{tabular}
\end{adjustbox}
\caption{We compute $p$-value by conducting a paired $t$-test between our approach and baselines with 100 repeated experiments for each metric on Adult dataset.}
\label{tab:p3}
\end{table*}

\begin{figure*}[!htb]
\centering
\includegraphics[width=1.0\textwidth]{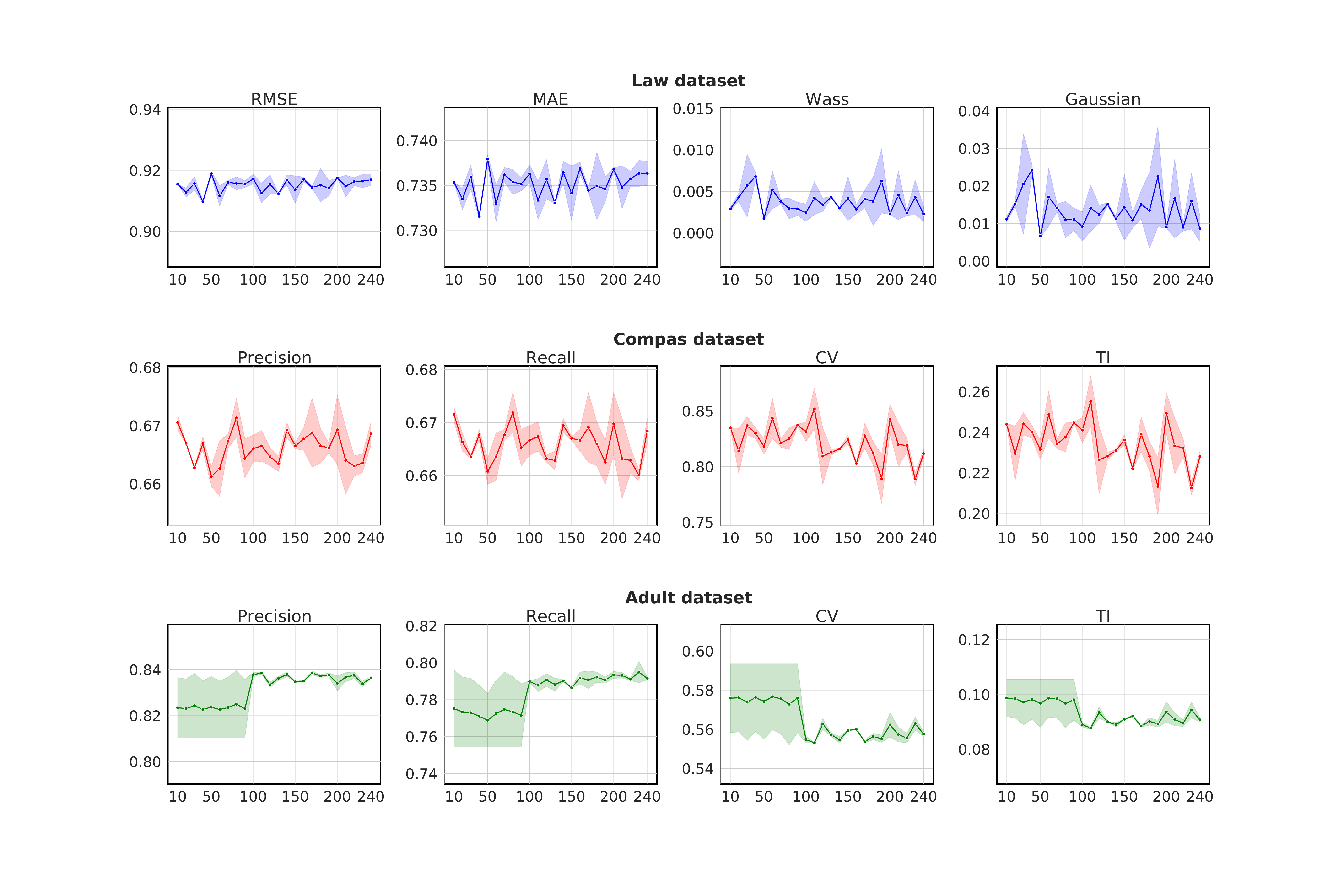}
\caption{We report the performance of our approach with different hyperparameter $\lambda$ on \texttt{Law}, \texttt{Compas} and \texttt{Adult} datasets. For each $\lambda$, we repeat the experiment 100 times to get the mean and variance.}
\label{fig:lambda}
\end{figure*}

\subsection{Implementation details}
All implementations are conducted in Python 3.7.7 with 64-bit Red Hat, Intel(R) Xeon(R) Gold 6150 CPU @ 2.70GHz.
The models for all datasets were trained with the following settings: 200 epochs, batch size of 64, Adam optimizer with the learning rate of $10^{-3}$, smooth loss function \cite{girshick2015fast} for \texttt{Law} dataset and cross-entropy loss function for \texttt{Adult} and \texttt{Compas} dataset. We used LeakyReLu \cite{maas2013rectifier} as the $h(.)$ function. We implemented the baseline methods by using Pyro library \cite{bingham2019pyro}, while our method was implemented by Pytorch. 
As regards the evaluation metric, we utilized the available functions from library AI360 \cite{aif360-oct-2018} and GeomLoss \cite{feydy2019interpolating}. More details of implementation settings can be found in the provided source code.

\subsection{Comparison results}


In this section, we report the empirical performance of different methods across three datasets on both the regression and classification tasks. In general, we aim to investigate the following research questions 1) how our approach achieves better fairness and accuracy tradeoff compared to other baselines; 2) how the model performance fluctuates with different hyperparameter $\lambda$ (the values of $\lambda$ for competitive and stable performance). 


\textbf{Regression task.} Table~\ref{table:law} indicates  the performance comparison for \texttt{Law} dataset. In particular, Full-LR and AE-LR models result in the best accuracy outcome with the lowest RMSE and MAE; however, this model fails to produce the fair prediction demonstrated by the highest fairness metrics. The possible reason is that both Full-LR and AE-LR use all features including the sensitive features, which is beneficial for the accuracy aspect but contains bias. The counterfactual fairness (CF$_1$- and CF$_2$-) and Multi-world methods in contrast witness a good performance when they come to fairness with a significantly low Wasserstein and Gaussian distance, but have quite high regression metrics. Meanwhile, our proposed method (InvFair) consistently produces the lowest results in Wasserstein and Gaussian distance and achieves quite competitive results in RMSE and MAE. We also observe that Linear Regression (-LR) performs better than Gradient Boosting Regression (-GBoostR). Finally, we notice that although the outstanding results are also recorded with Unaware-GBoostR in fairness aspects (0.009 for Wasserstein and 0.03 for Gaussian), its R2score is a negative number which implies the poor performance in the regression task.


\textbf{Classification task.} We analyze the task of classification on \texttt{Compas} and \texttt{Adult} datasets on Table ~\ref{table:compas} and Table~\ref{table:adult}, respectively. It is illustrated from Table~\ref{table:compas} that the poor performance is recorded with counterfactual fairness (CF$_1$- and CF$_2$-) and Multi-world approach with low results for classification metrics. In contrast, the latent representation produced from InvEnc combined with Logistic Regression and GBooost achieves the greatest results in terms of the classification metric including Balanced Accuracy, Precision, Recall and f-measure. Meanwhile, our proposed method surpasses all of the other methods regarding fairness metrics (CV and TI). It is moreover ranked second regarding Balanced Accuracy and F$_1$ score and ranked third regarding Precision and Recall. On the other hand, Table~\ref{table:adult} shows the results of different methods in \texttt{Adult} dataset. This dataset is highly imbalanced with the ratio of positive and negative classes being 70\% and 30\%. Our proposed approach produces the best Balanced Accuracy and Precision, while Full-GBoostC has the greatest F$_1$ and Recall score. Regarding fairness metrics, our method consistently surpasses all of the remaining methods. Moreover, gradient boosting classification (-GBoostC) performs better Logistic Regression model (-Log). As seen from the classification task, counterfactual fairness (CF$_1$- and CF$_2$-) and Multi-world model perform poorly in the classification task, possibly due to the misspecification of structural causal models. The invariant-encoder model (InvEnc) that minimizes the prediction of sensitive-awareness and fair-learning models allows the latent representation to achieve favorable outcomes in terms of accuracy aspects. Furthermore, when combined with the fair-learning models in our final approach (InvFair), it produces competitive results in both the prediction and fairness performance.



\textbf{Statistical significance.} To better comprehend the effectiveness of our proposed method in producing counterfactual samples compared with other approaches, we also perform a statistical significance test (paired $t$-test)
between our approach and other methods on each dataset and each metric with the obtained results on 100 randomly repeated experiments and report the result of $p$-value in Table~\ref{tab:p1}, Table~\ref{tab:p2} and Table~\ref{tab:p3}. We find that our model is statistically significant with $p < 0.05$, thus demonstrating the effectiveness of our proposed method in achieving counterfactual fairness.

\textbf{Sensitivity of hyperparameter.}
Figure \ref{fig:lambda} shows the variation of our proposed method performance with different settings of hyperparameter $\lambda$. For \texttt{Law} dataset, Gaussian distance fluctuates slightly from 0.01 to 0.02, while RMSE, MAE and Wasserstein are recorded at steady results. In terms of our proposed method performance on \texttt{Compas} and \texttt{Adult} datasets. In general, Precision and Recall share the same patterns, while CV and TI demonstrate similar trends. For \texttt{Compas} dataset, the performance of Precision and Recall have slight fluctuations of 0.66 and 0.68, while CV and TI vary marginally around 0.8-0.85 and 0.2-0.25, respectively. For \texttt{Adult} dataset, the performance witnesses a quite big variation before $\lambda$ reaches $100$, and thereafter achieves the outstanding and stable performance when $\lambda$ is greater than $100$.



\section{Conclusion}
This paper proposes a minimax game-theoretic approach that can maintain competitive performance in predictive tasks and make counterfactually fair decisions at the individual level.
We believe that training minimax objective functions for invariant-encoder model and fair-learning predictive model allow us
to exclude the sensitive information in models' decisions, and also maintain high accuracy performance. Empirical results on three real-world datasets demonstrated that our proposed approach (InvFair) performs best regarding fairness metrics and also achieves a favorable fairness-accuracy tradeoff. Most importantly, our approach does not require prior knowledge about the structural causal model, making it attractive in real-world applications. In future work, we plan to investigate how to estimate fair causal effects.

\bibliography{aaai}

\begin{thebibliography}{40}
\expandafter\ifx\csname natexlab\endcsname\relax\def\natexlab#1{#1}\fi
\providecommand{\url}[1]{\texttt{#1}}
\providecommand{\href}[2]{#2}
\providecommand{\path}[1]{#1}
\providecommand{\DOIprefix}{doi:}
\providecommand{\ArXivprefix}{arXiv:}
\providecommand{\URLprefix}{URL: }
\providecommand{\Pubmedprefix}{pmid:}
\providecommand{\doi}[1]{\href{http://dx.doi.org/#1}{\path{#1}}}
\providecommand{\Pubmed}[1]{\href{pmid:#1}{\path{#1}}}
\providecommand{\bibinfo}[2]{#2}
\ifx\xfnm\relax \def\xfnm[#1]{\unskip,\space#1}\fi
\bibitem[{Angwin et~al.(2016)Angwin, Larson, Mattu \&
  Kirchner}]{angwin2016machine}
\bibinfo{author}{Angwin, J.}, \bibinfo{author}{Larson, J.},
  \bibinfo{author}{Mattu, S.}, \& \bibinfo{author}{Kirchner, L.}
  (\bibinfo{year}{2016}).
\newblock \bibinfo{title}{Machine bias: There’s software used across the
  country to predict future criminals. and it’s biased against blacks.
  propublica (2016)}.
\newblock {\it \bibinfo{journal}{Google Scholar}\/},  (p.~\bibinfo{pages}{23}).
\bibitem[{Bellamy et~al.(2018)Bellamy, Dey, Hind, Hoffman, Houde, Kannan,
  Lohia, Martino, Mehta, Mojsilovic, Nagar, Ramamurthy, Richards, Saha,
  Sattigeri, Singh, Varshney \& Zhang}]{aif360-oct-2018}
\bibinfo{author}{Bellamy, R. K.~E.}, \bibinfo{author}{Dey, K.},
  \bibinfo{author}{Hind, M.}, \bibinfo{author}{Hoffman, S.~C.},
  \bibinfo{author}{Houde, S.}, \bibinfo{author}{Kannan, K.},
  \bibinfo{author}{Lohia, P.}, \bibinfo{author}{Martino, J.},
  \bibinfo{author}{Mehta, S.}, \bibinfo{author}{Mojsilovic, A.},
  \bibinfo{author}{Nagar, S.}, \bibinfo{author}{Ramamurthy, K.~N.},
  \bibinfo{author}{Richards, J.}, \bibinfo{author}{Saha, D.},
  \bibinfo{author}{Sattigeri, P.}, \bibinfo{author}{Singh, M.},
  \bibinfo{author}{Varshney, K.~R.}, \& \bibinfo{author}{Zhang, Y.}
  (\bibinfo{year}{2018}).
\newblock \bibinfo{title}{{AI Fairness} 360: An extensible toolkit for
  detecting, understanding, and mitigating unwanted algorithmic bias}.
\newblock \URLprefix \url{https://arxiv.org/abs/1810.01943}.
\bibitem[{Berk et~al.(2021)Berk, Heidari, Jabbari, Kearns \&
  Roth}]{berk2021fairness}
\bibinfo{author}{Berk, R.}, \bibinfo{author}{Heidari, H.},
  \bibinfo{author}{Jabbari, S.}, \bibinfo{author}{Kearns, M.}, \&
  \bibinfo{author}{Roth, A.} (\bibinfo{year}{2021}).
\newblock \bibinfo{title}{Fairness in criminal justice risk assessments: The
  state of the art}.
\newblock {\it \bibinfo{journal}{Sociological Methods \& Research}\/},  {\it
  \bibinfo{volume}{50}\/}, \bibinfo{pages}{3--44}.
\bibitem[{Biega et~al.(2018)Biega, Gummadi \& Weikum}]{biega2018equity}
\bibinfo{author}{Biega, A.~J.}, \bibinfo{author}{Gummadi, K.~P.}, \&
  \bibinfo{author}{Weikum, G.} (\bibinfo{year}{2018}).
\newblock \bibinfo{title}{Equity of attention: Amortizing individual fairness
  in rankings}.
\newblock In {\it \bibinfo{booktitle}{The 41st international acm sigir
  conference on research \& development in information retrieval}\/} (pp.
  \bibinfo{pages}{405--414}).
\bibitem[{Bingham et~al.(2019)Bingham, Chen, Jankowiak, Obermeyer, Pradhan,
  Karaletsos, Singh, Szerlip, Horsfall \& Goodman}]{bingham2019pyro}
\bibinfo{author}{Bingham, E.}, \bibinfo{author}{Chen, J.~P.},
  \bibinfo{author}{Jankowiak, M.}, \bibinfo{author}{Obermeyer, F.},
  \bibinfo{author}{Pradhan, N.}, \bibinfo{author}{Karaletsos, T.},
  \bibinfo{author}{Singh, R.}, \bibinfo{author}{Szerlip, P.},
  \bibinfo{author}{Horsfall, P.}, \& \bibinfo{author}{Goodman, N.~D.}
  (\bibinfo{year}{2019}).
\newblock \bibinfo{title}{Pyro: Deep universal probabilistic programming}.
\newblock {\it \bibinfo{journal}{The Journal of Machine Learning Research}\/},
  {\it \bibinfo{volume}{20}\/}, \bibinfo{pages}{973--978}.
\bibitem[{Bollen \& Pearl(2013)}]{bollen2013eight}
\bibinfo{author}{Bollen, K.~A.}, \& \bibinfo{author}{Pearl, J.}
  (\bibinfo{year}{2013}).
\newblock \bibinfo{title}{Eight myths about causality and structural equation
  models}.
\newblock In {\it \bibinfo{booktitle}{Handbook of causal analysis for social
  research}\/} (pp. \bibinfo{pages}{301--328}).
\newblock \bibinfo{publisher}{Springer}.
\bibitem[{Brodersen et~al.(2010)Brodersen, Ong, Stephan \&
  Buhmann}]{brodersen2010balanced}
\bibinfo{author}{Brodersen, K.~H.}, \bibinfo{author}{Ong, C.~S.},
  \bibinfo{author}{Stephan, K.~E.}, \& \bibinfo{author}{Buhmann, J.~M.}
  (\bibinfo{year}{2010}).
\newblock \bibinfo{title}{The balanced accuracy and its posterior
  distribution}.
\newblock In {\it \bibinfo{booktitle}{2010 20th international conference on
  pattern recognition}\/} (pp. \bibinfo{pages}{3121--3124}).
\newblock \bibinfo{organization}{IEEE}.
\bibitem[{Chen et~al.(2019)Chen, Kallus, Mao, Svacha \&
  Udell}]{chen2019fairness}
\bibinfo{author}{Chen, J.}, \bibinfo{author}{Kallus, N.}, \bibinfo{author}{Mao,
  X.}, \bibinfo{author}{Svacha, G.}, \& \bibinfo{author}{Udell, M.}
  (\bibinfo{year}{2019}).
\newblock \bibinfo{title}{Fairness under unawareness: Assessing disparity when
  protected class is unobserved}.
\newblock In {\it \bibinfo{booktitle}{Proceedings of the conference on
  fairness, accountability, and transparency}\/} (pp.
  \bibinfo{pages}{339--348}).
\bibitem[{Chiappa(2019)}]{chiappa2019path}
\bibinfo{author}{Chiappa, S.} (\bibinfo{year}{2019}).
\newblock \bibinfo{title}{Path-specific counterfactual fairness}.
\newblock In {\it \bibinfo{booktitle}{Proceedings of the AAAI Conference on
  Artificial Intelligence}\/} (pp. \bibinfo{pages}{7801--7808}).
\newblock volume~\bibinfo{volume}{33}.
\bibitem[{Cover et~al.(1991)Cover, Thomas et~al.}]{cover1991entropy}
\bibinfo{author}{Cover, T.~M.}, \bibinfo{author}{Thomas, J.~A.} et~al.
  (\bibinfo{year}{1991}).
\newblock \bibinfo{title}{Entropy, relative entropy and mutual information}.
\newblock {\it \bibinfo{journal}{Elements of information theory}\/},  {\it
  \bibinfo{volume}{2}\/}, \bibinfo{pages}{12--13}.
\bibitem[{Dua \& Graff(2017)}]{Dua:2019}
\bibinfo{author}{Dua, D.}, \& \bibinfo{author}{Graff, C.}
  (\bibinfo{year}{2017}).
\newblock \bibinfo{title}{{UCI} machine learning repository}.
\newblock \URLprefix \url{http://archive.ics.uci.edu/ml}.
\bibitem[{Dwork et~al.(2012)Dwork, Hardt, Pitassi, Reingold \&
  Zemel}]{dwork2012fairness}
\bibinfo{author}{Dwork, C.}, \bibinfo{author}{Hardt, M.},
  \bibinfo{author}{Pitassi, T.}, \bibinfo{author}{Reingold, O.}, \&
  \bibinfo{author}{Zemel, R.} (\bibinfo{year}{2012}).
\newblock \bibinfo{title}{Fairness through awareness}.
\newblock In {\it \bibinfo{booktitle}{Proceedings of the 3rd innovations in
  theoretical computer science conference}\/} (pp. \bibinfo{pages}{214--226}).
\bibitem[{Feydy et~al.(2019)Feydy, S{\'e}journ{\'e}, Vialard, Amari, Trouve \&
  Peyr{\'e}}]{feydy2019interpolating}
\bibinfo{author}{Feydy, J.}, \bibinfo{author}{S{\'e}journ{\'e}, T.},
  \bibinfo{author}{Vialard, F.-X.}, \bibinfo{author}{Amari, S.-i.},
  \bibinfo{author}{Trouve, A.}, \& \bibinfo{author}{Peyr{\'e}, G.}
  (\bibinfo{year}{2019}).
\newblock \bibinfo{title}{Interpolating between optimal transport and mmd using
  sinkhorn divergences}.
\newblock In {\it \bibinfo{booktitle}{The 22nd International Conference on
  Artificial Intelligence and Statistics}\/} (pp. \bibinfo{pages}{2681--2690}).
\bibitem[{Fong(2013)}]{fong2013causal}
\bibinfo{author}{Fong, B.} (\bibinfo{year}{2013}).
\newblock \bibinfo{title}{Causal theories: A categorical perspective on
  bayesian networks}.
\newblock {\it \bibinfo{journal}{arXiv preprint arXiv:1301.6201}\/}, .
\bibitem[{Girshick(2015)}]{girshick2015fast}
\bibinfo{author}{Girshick, R.} (\bibinfo{year}{2015}).
\newblock \bibinfo{title}{Fast r-cnn}.
\newblock In {\it \bibinfo{booktitle}{Proceedings of the IEEE international
  conference on computer vision}\/} (pp. \bibinfo{pages}{1440--1448}).
\bibitem[{Gretton et~al.(2012)Gretton, Borgwardt, Rasch, Sch{\"o}lkopf \&
  Smola}]{gretton2012kernel}
\bibinfo{author}{Gretton, A.}, \bibinfo{author}{Borgwardt, K.~M.},
  \bibinfo{author}{Rasch, M.~J.}, \bibinfo{author}{Sch{\"o}lkopf, B.}, \&
  \bibinfo{author}{Smola, A.} (\bibinfo{year}{2012}).
\newblock \bibinfo{title}{A kernel two-sample test}.
\newblock {\it \bibinfo{journal}{The Journal of Machine Learning Research}\/},
  {\it \bibinfo{volume}{13}\/}, \bibinfo{pages}{723--773}.
\bibitem[{Grgic-Hlaca et~al.(2016)Grgic-Hlaca, Zafar, Gummadi \&
  Weller}]{grgic2016case}
\bibinfo{author}{Grgic-Hlaca, N.}, \bibinfo{author}{Zafar, M.~B.},
  \bibinfo{author}{Gummadi, K.~P.}, \& \bibinfo{author}{Weller, A.}
  (\bibinfo{year}{2016}).
\newblock \bibinfo{title}{The case for process fairness in learning: Feature
  selection for fair decision making}.
\newblock In {\it \bibinfo{booktitle}{NIPS Symposium on Machine Learning and
  the Law}\/} (p.~\bibinfo{pages}{2}).
\newblock volume~\bibinfo{volume}{1}.
\bibitem[{Kingma \& Ba(2014)}]{kingma2014adam}
\bibinfo{author}{Kingma, D.~P.}, \& \bibinfo{author}{Ba, J.}
  (\bibinfo{year}{2014}).
\newblock \bibinfo{title}{Adam: A method for stochastic optimization}.
\newblock {\it \bibinfo{journal}{arXiv preprint arXiv:1412.6980}\/}, .
\bibitem[{Kusner et~al.(2017)Kusner, Loftus, Russell \&
  Silva}]{kusner2017counterfactual}
\bibinfo{author}{Kusner, M.~J.}, \bibinfo{author}{Loftus, J.~R.},
  \bibinfo{author}{Russell, C.}, \& \bibinfo{author}{Silva, R.}
  (\bibinfo{year}{2017}).
\newblock \bibinfo{title}{Counterfactual fairness}.
\newblock {\it \bibinfo{journal}{arXiv preprint arXiv:1703.06856}\/}, .
\bibitem[{Larson et~al.(2016)Larson, Mattu, Kirchner \& Angwin}]{larson2016we}
\bibinfo{author}{Larson, J.}, \bibinfo{author}{Mattu, S.},
  \bibinfo{author}{Kirchner, L.}, \& \bibinfo{author}{Angwin, J.}
  (\bibinfo{year}{2016}).
\newblock \bibinfo{title}{How we analyzed the compas recidivism algorithm}.
\newblock {\it \bibinfo{journal}{ProPublica (5 2016)}\/},  {\it
  \bibinfo{volume}{9}\/}.
\bibitem[{Maas et~al.(2013)Maas, Hannun, Ng et~al.}]{maas2013rectifier}
\bibinfo{author}{Maas, A.~L.}, \bibinfo{author}{Hannun, A.~Y.},
  \bibinfo{author}{Ng, A.~Y.} et~al. (\bibinfo{year}{2013}).
\newblock \bibinfo{title}{Rectifier nonlinearities improve neural network
  acoustic models}.
\newblock In {\it \bibinfo{booktitle}{Proc. icml}\/} (p.~\bibinfo{pages}{3}).
\newblock \bibinfo{organization}{Citeseer} volume~\bibinfo{volume}{30}.
\bibitem[{McDiarmid et~al.(1989)}]{mcdiarmid1989method}
\bibinfo{author}{McDiarmid, C.} et~al. (\bibinfo{year}{1989}).
\newblock \bibinfo{title}{On the method of bounded differences}.
\newblock {\it \bibinfo{journal}{Surveys in combinatorics}\/},  {\it
  \bibinfo{volume}{141}\/}, \bibinfo{pages}{148--188}.
\bibitem[{Miconi(2017)}]{miconi2017impossibility}
\bibinfo{author}{Miconi, T.} (\bibinfo{year}{2017}).
\newblock \bibinfo{title}{The impossibility of" fairness": a generalized
  impossibility result for decisions}.
\newblock {\it \bibinfo{journal}{arXiv preprint arXiv:1707.01195}\/}, .
\bibitem[{Mukherjee et~al.(2020)Mukherjee, Yurochkin, Banerjee \&
  Sun}]{mukherjee2020two}
\bibinfo{author}{Mukherjee, D.}, \bibinfo{author}{Yurochkin, M.},
  \bibinfo{author}{Banerjee, M.}, \& \bibinfo{author}{Sun, Y.}
  (\bibinfo{year}{2020}).
\newblock \bibinfo{title}{Two simple ways to learn individual fairness metrics
  from data}.
\newblock In {\it \bibinfo{booktitle}{International Conference on Machine
  Learning}\/} (pp. \bibinfo{pages}{7097--7107}).
\newblock \bibinfo{organization}{PMLR}.
\bibitem[{Nabi \& Shpitser(2018)}]{nabi2018fair}
\bibinfo{author}{Nabi, R.}, \& \bibinfo{author}{Shpitser, I.}
  (\bibinfo{year}{2018}).
\newblock \bibinfo{title}{Fair inference on outcomes}.
\newblock In {\it \bibinfo{booktitle}{Proceedings of the AAAI Conference on
  Artificial Intelligence}\/}.
\newblock volume~\bibinfo{volume}{32}.
\bibitem[{Ng et~al.(2011)}]{ng2011sparse}
\bibinfo{author}{Ng, A.} et~al. (\bibinfo{year}{2011}).
\newblock \bibinfo{title}{Sparse autoencoder}.
\newblock {\it \bibinfo{journal}{CS294A Lecture notes}\/},  {\it
  \bibinfo{volume}{72}\/}, \bibinfo{pages}{1--19}.
\bibitem[{Oh et~al.(2019)Oh, Pouryahya, Iyer, Apte, Tannenbaum \&
  Deasy}]{oh2019kernel}
\bibinfo{author}{Oh, J.~H.}, \bibinfo{author}{Pouryahya, M.},
  \bibinfo{author}{Iyer, A.}, \bibinfo{author}{Apte, A.~P.},
  \bibinfo{author}{Tannenbaum, A.}, \& \bibinfo{author}{Deasy, J.~O.}
  (\bibinfo{year}{2019}).
\newblock \bibinfo{title}{Kernel wasserstein distance}.
\newblock {\it \bibinfo{journal}{arXiv preprint arXiv:1905.09314}\/}, .
\bibitem[{Pearl(2009{\natexlab{a}})}]{pearl2009causal}
\bibinfo{author}{Pearl, J.} (\bibinfo{year}{2009}{\natexlab{a}}).
\newblock \bibinfo{title}{Causal inference in statistics: An overview}.
\newblock {\it \bibinfo{journal}{Statistics surveys}\/},  {\it
  \bibinfo{volume}{3}\/}, \bibinfo{pages}{96--146}.
\bibitem[{Pearl(2009{\natexlab{b}})}]{pearl2009causality}
\bibinfo{author}{Pearl, J.} (\bibinfo{year}{2009}{\natexlab{b}}).
\newblock {\it \bibinfo{title}{Causality}\/}.
\newblock \bibinfo{publisher}{Cambridge university press}.
\bibitem[{Pearl(2012)}]{pearl2012causal}
\bibinfo{author}{Pearl, J.} (\bibinfo{year}{2012}).
\newblock {\it \bibinfo{title}{The causal foundations of structural equation
  modeling}\/}.
\newblock \bibinfo{type}{Technical Report} CALIFORNIA UNIV LOS ANGELES DEPT OF
  COMPUTER SCIENCE.
\bibitem[{Peters et~al.(2016)Peters, B{\"u}hlmann \&
  Meinshausen}]{peters2016causal}
\bibinfo{author}{Peters, J.}, \bibinfo{author}{B{\"u}hlmann, P.}, \&
  \bibinfo{author}{Meinshausen, N.} (\bibinfo{year}{2016}).
\newblock \bibinfo{title}{Causal inference by using invariant prediction:
  identification and confidence intervals}.
\newblock {\it \bibinfo{journal}{Journal of the Royal Statistical Society.
  Series B (Statistical Methodology)}\/},  (pp. \bibinfo{pages}{947--1012}).
\bibitem[{R{\"u}schendorf(1985)}]{ruschendorf1985wasserstein}
\bibinfo{author}{R{\"u}schendorf, L.} (\bibinfo{year}{1985}).
\newblock \bibinfo{title}{The wasserstein distance and approximation theorems}.
\newblock {\it \bibinfo{journal}{Probability Theory and Related Fields}\/},
  {\it \bibinfo{volume}{70}\/}, \bibinfo{pages}{117--129}.
\bibitem[{Russell et~al.(2017)Russell, Kusner, Loftus \&
  Silva}]{russell2017worlds}
\bibinfo{author}{Russell, C.}, \bibinfo{author}{Kusner, M.~J.},
  \bibinfo{author}{Loftus, J.~R.}, \& \bibinfo{author}{Silva, R.}
  (\bibinfo{year}{2017}).
\newblock \bibinfo{title}{When worlds collide: integrating different
  counterfactual assumptions in fairness}.
\newblock {\it \bibinfo{journal}{Advances in Neural Information Processing
  Systems 30. Pre-proceedings}\/},  {\it \bibinfo{volume}{30}\/}.
\bibitem[{Sharifi-Malvajerdi et~al.(2019)Sharifi-Malvajerdi, Kearns \&
  Roth}]{sharifi2019average}
\bibinfo{author}{Sharifi-Malvajerdi, S.}, \bibinfo{author}{Kearns, M.}, \&
  \bibinfo{author}{Roth, A.} (\bibinfo{year}{2019}).
\newblock \bibinfo{title}{Average individual fairness: Algorithms,
  generalization and experiments}.
\newblock {\it \bibinfo{journal}{Advances in Neural Information Processing
  Systems}\/},  {\it \bibinfo{volume}{32}\/}, \bibinfo{pages}{8242--8251}.
\bibitem[{Speicher et~al.(2018)Speicher, Heidari, Grgic-Hlaca, Gummadi, Singla,
  Weller \& Zafar}]{speicher2018unified}
\bibinfo{author}{Speicher, T.}, \bibinfo{author}{Heidari, H.},
  \bibinfo{author}{Grgic-Hlaca, N.}, \bibinfo{author}{Gummadi, K.~P.},
  \bibinfo{author}{Singla, A.}, \bibinfo{author}{Weller, A.}, \&
  \bibinfo{author}{Zafar, M.~B.} (\bibinfo{year}{2018}).
\newblock \bibinfo{title}{A unified approach to quantifying algorithmic
  unfairness: Measuring individual \&group unfairness via inequality indices}.
\newblock In {\it \bibinfo{booktitle}{Proceedings of the 24th ACM SIGKDD
  International Conference on Knowledge Discovery \& Data Mining}\/} (pp.
  \bibinfo{pages}{2239--2248}).
\bibitem[{VanderWeele(2009)}]{vanderweele2009concerning}
\bibinfo{author}{VanderWeele, T.~J.} (\bibinfo{year}{2009}).
\newblock \bibinfo{title}{Concerning the consistency assumption in causal
  inference}.
\newblock {\it \bibinfo{journal}{Epidemiology}\/},  {\it
  \bibinfo{volume}{20}\/}, \bibinfo{pages}{880--883}.
\bibitem[{Wightman(1998)}]{wightman1998lsac}
\bibinfo{author}{Wightman, L.~F.} (\bibinfo{year}{1998}).
\newblock \bibinfo{title}{Lsac national longitudinal bar passage study. lsac
  research report series.}, .
\bibitem[{Wu et~al.(2019)Wu, Zhang \& Wu}]{wu2019counterfactual}
\bibinfo{author}{Wu, Y.}, \bibinfo{author}{Zhang, L.}, \& \bibinfo{author}{Wu,
  X.} (\bibinfo{year}{2019}).
\newblock \bibinfo{title}{Counterfactual fairness: Unidentification, bound and
  algorithm}.
\newblock In {\it \bibinfo{booktitle}{Proceedings of the Twenty-Eighth
  International Joint Conference on Artificial Intelligence}\/}.
\bibitem[{Zhang \& Bareinboim(2018)}]{zhang2018fairness}
\bibinfo{author}{Zhang, J.}, \& \bibinfo{author}{Bareinboim, E.}
  (\bibinfo{year}{2018}).
\newblock \bibinfo{title}{Fairness in decision-making—the causal explanation
  formula}.
\newblock In {\it \bibinfo{booktitle}{Thirty-Second AAAI Conference on
  Artificial Intelligence}\/}.
\bibitem[{Zhang \& Zhou(2019)}]{zhang2019fairness}
\bibinfo{author}{Zhang, Y.}, \& \bibinfo{author}{Zhou, L.}
  (\bibinfo{year}{2019}).
\newblock \bibinfo{title}{Fairness assessment for artificial intelligence in
  financial industry}.
\newblock {\it \bibinfo{journal}{arXiv preprint arXiv:1912.07211}\/}, .

\end{thebibliography}

\end{document}